\documentclass{article}

\usepackage[nonatbib,final]{neurips_2024}

\usepackage[utf8]{inputenc} 
\usepackage[T1]{fontenc} 
\usepackage{hyperref}

\usepackage{url}            
\usepackage{booktabs}       
\usepackage{amsmath}
\usepackage{cleveref}
\usepackage{amsfonts}       
\usepackage{nicefrac}       
\usepackage{microtype}      
\usepackage{comment}
\usepackage{tabularx}
\usepackage{graphicx}
\usepackage{pifont}
\graphicspath{{figures/}{figures/}}
\usepackage{caption}
\usepackage{wrapfig}
\usepackage[table]{xcolor} 
\usepackage[ruled,vlined]{algorithm2e}
\newcolumntype{Y}{>{\centering\arraybackslash}X}
\usepackage{array,multirow}
\usepackage{colortbl}
\usepackage{enumitem}
\usepackage{wrapfig} 

\usepackage{subcaption}
\usepackage{tikz}
\tikzset{fontscale/.style = {font=\relsize{#1}}
    }

\crefname{figure}{Figure}{Figures}
\Crefname{figure}{Figure}{Figures}
\crefname{table}{Table}{Tables}
\Crefname{table}{Table}{Tables}
\crefname{equation}{Eq.}{Eqs.}
\Crefname{equation}{Eq.}{Eqs.}
\crefname{appendix}{Appendix}{Appendix}
\Crefname{appendix}{Appendix}{Appendix}
\SetCommentSty{mycommfont}
\definecolor{darkgreen}{rgb}{0.0, 0.5, 0.0}
\definecolor{darkyellow}{rgb}{0.8, 0.8, 0.0} 

\definecolor{slatecolor}{rgb}{1, 0.95, 0.9}

\newcommand{\minus}{\scalebox{0.5}[0.9]{$-$}}

\hypersetup{
    colorlinks = true,
    linkcolor = {red},
    citecolor = {darkgreen},
    urlcolor = {darkgreen},
    allbordercolors = {white}
} 
\newcommand{\cmark}{\ding{51}}%
\newcommand{\xmark}{\ding{55}}%
\newcommand{\eg}{\textit{e.g.} }

\newcommand{\link}{\texttt{Edge}\xspace}
\usepackage{xspace}

\newcommand{\ours}{\text{SLATE}\xspace}

\title{Supra-Laplacian Encoding for \\ Transformer on Dynamic Graphs}

\author{%
  Yannis Karmim \\
  Conservatoire National des Arts et Métiers\\
  CEDRIC, EA 4629 \\
  F 75003, Paris, France \\
  \texttt{yannis.karmim@cnam.fr} \\
  \And
  Marc Lafon\\
  Conservatoire National des Arts et Métiers\\
  CEDRIC, EA 4629 \\
  F 75003, Paris, France \\
  \texttt{marc.lafon@lecnam.net} \\
  \AND
Raphaël Fournier S'niehotta \\
   Conservatoire National des Arts et Métiers\\
   CEDRIC, EA 4629 \\
   F 75003, Paris, France\\
  \texttt{fournier@cnam.fr} \\
  \And
  Nicolas Thome \\
  Sorbonne Université \\
  CNRS, ISIR \\
  F-75005 Paris, France\\
  \texttt{nicolas.thome@isir.upmc.fr} \\
}

\begin{document}

\maketitle


\begin{abstract}
Fully connected Graph Transformers (GT) have rapidly become prominent in the static graph community as an alternative to Message-Passing models, which suffer from a lack of expressivity, oversquashing, and under-reaching.
However, in a dynamic context, by interconnecting all nodes at multiple snapshots with self-attention,GT loose both structural and temporal information. In this work, we introduce \textbf{S}upra-\textbf{LA}placian encoding for spatio-temporal \textbf{T}ransform\textbf{E}rs (\ours), a new spatio-temporal encoding to leverage the GT architecture while keeping spatio-temporal information.
Specifically, we transform Discrete Time Dynamic Graphs into multi-layer graphs and take advantage of the spectral properties of their associated supra-Laplacian matrix.
Our second contribution explicitly model nodes' pairwise relationships with a cross-attention mechanism, providing an accurate edge representation for dynamic link prediction.
\ours outperforms numerous state-of-the-art methods based on Message-Passing Graph Neural Networks combined with recurrent models (\eg, LSTM), and Dynamic Graph Transformers,
on~9 datasets. Code is open-source and available at this link \href{https://github.com/ykrmm/SLATE}{https://github.com/ykrmm/SLATE}.
\end{abstract}



\section{Introduction}

Dynamic graphs are crucial for modeling interactions between entities in various fields, from social sciences to computational biology \cite{ying2018graphpinsage, he2020lightgcn:recommendation, jumper2021highlyalphafold, kaba2022equivariant}. Link prediction on dynamic graphs is an all-important task, with diverse applications, such as predicting user actions in recommender systems, forecasting financial transactions, or identifying potential academic collaborations. Dynamic graphs can be modeled as a time series of static graphs captured at regular intervals (Discrete Time Dynamic Graphs, DTDG) \cite{skarding2021foundationssurvey, yang2024dynamicsurvey}.

 Standard approaches for learning representations on DTDGs combine Message-Passing GNNs (MP-GNNs) with temporal RNN-based models \cite{you2022roland:graphs, sankar2018dysat:networks, pareja2019evolvegcn:graphs}. In static contexts, Graph Transformers (GT) \cite{dwivedi2020agraphs, wu2023simplifyingrepresentations, kreuzer2021rethinkingattention} offer a compelling alternative to MP-GNNs that faced several limitations \cite{xu2018hownetworks,topping2021understandingcurvature}. Indeed, their fully-connected attention mechanism captures long-range dependencies, resolving issues such as oversquashing \cite{alon2020onimplications}. GTs directly connect nodes, using the graph structure as a soft bias through positional encoding \cite{rampasek2022recipetransformer}. Incorporating Laplacian-based encodings in GTs provably enhances their expressiveness compared to MP-GNNs \cite{kreuzer2021rethinkingattention, dwivedi2020agraphs}.

Exploiting GTs on dynamic graphs would require a spatio-temporal encoding that effectively retains both structural and temporal information. The recent works that have extended GTs to dynamic graphs capture spatio-temporal dependencies between nodes by using \textit{partial} attention mechanisms~\cite{liu2021anomalytransformer, yang2022time-awareevolution, wang2021tcl:learning, hu2023recurrentstates}. Moreover, these methods also employ encodings which \textit{independently} embed the graph structure and the temporal dimension. Given that the expressiveness of GTs depends on an accurate spatio-temporal encoding, designing one that interweaves time and position information could greatly enhance their potential and performance.

The vast majority of neural-based methods for dynamic link prediction rely on node representation learning \cite{pareja2019evolvegcn:graphs, yang2021discrete-timespace, rossi2020tgn:graphs, you2022roland:graphs, sankar2018dysat:networks}. Recent works enrich node embeddings with pairwise information for a given node-pair using co-occurrence neighbors matching \cite{yu2023towardslibrary,wang2021inductivewalks} or cross-attention on historical sub-graphs \cite{wang2021tcl:learning}. However these methods neglect the global information of the graph by sampling different spatio-temporal substructures around targeted nodes. 

\begin{figure}[t]
    \centering
    \includegraphics[width=14cm]{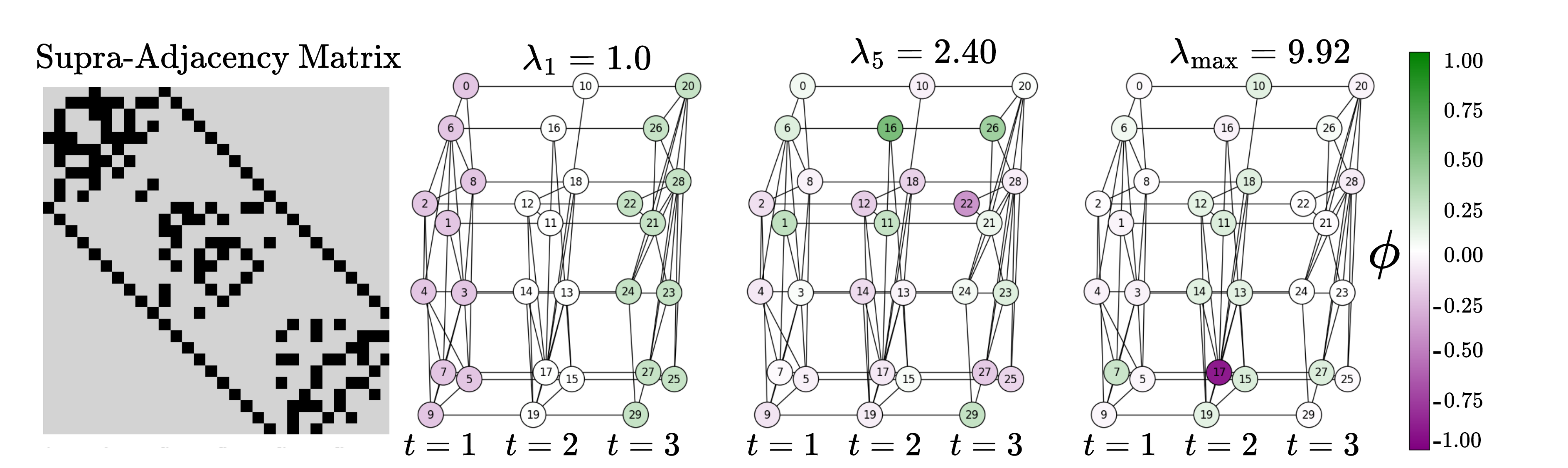}

    \caption{ \ours is a fully connected transformer for dynamic link prediction, which innovatively performs a joint spatial and temporal encoding of the dynamic graph. \ours models a DTDG as a multi-layer graph with temporal dependencies between a node and its past.
    Building the supra-adjacency matrix of a randomly-generated toy dynamic graph with 3 snapshots (\textit{left}) and analysing the spectrum of its associated supra-Laplacian (\textit{right}) provide fundamental spatio-temporal information. The projections on eigenvectors associated with smaller eigenvalues ($\lambda_1$) capture global graph dynamics: node colors are different for each time step. Larger eigenvalues ( \eg $\lambda_{\text{max}}$), capture more localized spatio-temporal information (see~\cref{app:rw_multilayer}).    
    }
    \label{fig:supralap}
\end{figure}

Pioneering work in the complex network community has studied temporal graphs
with multi-layers models and supra-adjacency matrices~\cite{valdano2015analyticalnetworks, kivela2014multilayer}. The spectral analysis of such matrices can provide valuable structural and temporal information~\cite{cozzomultilayerproperties, radicchi2013abruptnetworks}. However, how to adapt this formalism for learning dynamic graphs with transformer architectures remains a widely open question. 

In this work, we introduce \textbf{S}upra-\textbf{LA}placian encoding for spatio-temporal \textbf{T}ransform\textbf{E}rs (\ours), a new unified spatio-temporal encoding which allows to fully exploit the potential of the GT architecture for the task of dynamic link prediction. 
As illustrated on~\cref{fig:supralap}, adapting supra-Laplacian matrices to dynamic graph can provide rich spatio-temporal information for positional encoding. 
\ours is based on the following two main contributions:

\begin{itemize}

\item We bridge the gap between multi-layer networks and Discrete Time Dynamic Graphs (DTDGs) by adapting the spectral properties of supra-Laplacian matrices for transformers on dynamic graphs. By carefully transforming the supra-Laplacian matrices for DTDGs, we derive a connected multi-layer graph that captures various levels of spatio-temporal information. We introduce a fully-connected spatio-temporal transformer that leverages this unified supra-Laplacian encoding.

    \item The proposed transfomer captures dependencies between nodes across multiple time steps, creating dynamic representations. 

    To enhance link prediction, we introduce a lightweight edge representation module using \textit{cross-attention} only between the temporal representations of node pairs, precisely capturing their evolving interactions. This results in a unique edge embedding, significantly streamlining the prediction process and boosting both efficiency and accuracy.

\end{itemize}

We conduct an extensive experimental validation of our method across 11 real and synthetic discrete-time dynamic graph datasets. 
\ours outperforms state-of-the-art results by a large margin.
~We also validate the importance of our supra-Laplacian unified spatio-temporal encoding and the edge module for optimal performances. Finally, \ours remains efficient since it uses a single-layer transformer, and we show impressive results on larger graph datasets, indicating good scalability, and limited time-memory overhead.

\section{Related work}

\paragraph{Dynamic Graph Neural Networks on DTDGs.}
The standard approach to learn on DTDGs
\cite{skarding2021foundationssurvey, yang2024dynamicsurvey} involves using two
separate spatial and temporal models. The spatial model is responsible for
encoding the structure of the current graph snapshot, while the temporal model
updates the dynamic either of the graph representations
\cite{sankar2018dysat:networks, Seo2016StructuredNetworks,you2022roland:graphs,li2019predictinggraphs,kuo2023dynamictransformer} or the graph model parameters \cite{pareja2019evolvegcn:graphs,hajiramezanali2019variationalnetworks}. Recently, ROLAND \cite{you2022roland:graphs} introduced a generic framework to use any static graph model for spatial encoding coupled with a recurrent-based (LSTM \cite{hochreiter1997longmemory}, RNN, GRU) or attention-based temporal model.
These above methods mainly use a MP-GNN as spatial model~\cite{kipf2016semi-supervisednetworks,velickovic2017graphnetworks,ying2018graphpinsage}. However, MP-GNNs are known to present critical limitations: they struggle to distinguish simple structures like triangles or cycles \cite{morris2018weisfeilernetworks,chen2020cansubstructures}, and  fail to capture long-range dependencies due to oversquashing~\cite{alon2020onimplications, topping2021understandingcurvature}. To overcome these limitations, some works have adopted a fully-connected GT as spatial model, benefiting from its global attention mechanism~\cite{chutransmot:tracking,wei2022dgtr:detection,zheng2023vdgcnet:model}. In~\cite{sankar2018dysat:networks}, the local structure is preserved by computing the attention on direct neighbors. In contrast to these works, \ours uses a unique spatio-temporal graph transformer model, greatly simplifying the learning process.  

\paragraph{Graph Transformer.} In static contexts, Graph Transformers have been shown to provide a compelling alternative to MP-GNNs~\cite{dwivedi2020agraphs}. GTs~\cite{wu2023simplifyingrepresentations,rampasek2022recipetransformer,kim2022purelearners,ying2021graphormer:representation} enable direct connections between all nodes, using the graph’s structure as a soft inductive bias, thus resolving the oversquashing issue. The expressiveness of GTs heavily depends on positional or structural encoding \cite{dwivedi2020agraphs,mialon2021graphit:transformers,dwivedi2021graphrepresentations,beaini2020directionalnetworks}. In~\cite{dwivedi2020agraphs}, the authors use the eigenvectors associated with the $k$-lowest eigenvalues of the Laplacian matrix, which allows GTs to distinguish structures that MP-GNNs are unable to differentiate. Following the success of Laplacian positional encoding on static graphs, \ours uses the eigenvectors of the supra-Laplacian of a multi-layer graph representation of DTDGs as spatio-temporal encoding. 

\paragraph{Dynamic Graph Transformers.}
To avoid separately modelling structural and temporal information as dynamic Graph Neural Networks usually do on DTDGs, recent papers have adopted a unified model based on spatio-temporal attention \cite{liu2021anomalytransformer,hu2023recurrentstates}. This novel approach make those models close to transformer-based methods classically employed to learn on Continuous Time Dynamic Graphs (CTDG)~\cite{xu2020tgat:graphs,wang2021tcl:learning,wang2022temporalnetwork}. Among them, some preserve the local structure by computing attention only on direct neighbors~\cite{xu2020tgat:graphs,sankar2018dysat:networks}, while others sample local spatio-temporal structures around nodes~\cite{wang2021tcl:learning,liu2021anomalytransformer,yang2022time-awareevolution,hu2023recurrentstates} and perform fully-connected attention. 
However, their spatio-temporal encoding is still built by concatenating a spatial and a temporal encoding that are computed independently.  The spatial encoding is either based on a graph-based distance \cite{wang2021tcl:learning,hu2023recurrentstates} or on a diffusion-based measure~\cite{liu2021anomalytransformer}. The temporal encoding is usually sinus-based~\cite{xu2020tgat:graphs,wang2022temporalnetwork,banerjee2022spatial-temporalforecasting} as in the original transformer paper \cite{vaswani2017attentionneed}. Another drawback of these methods \cite{liu2021anomalytransformer,wang2021tcl:learning,hu2023recurrentstates,wang2022temporalnetwork} is that they use only sub-graphs to represent the local structure around a given node. Therefore, their representations of the nodes are computed on different graphs and thus fail to capture global and long-range interactions. Contrary to those approaches, our \ours model uses the same graph to compute node representations in a fully-connected GT between all nodes within temporal windows. It features a unified spatio-temporal encoding based on the supra-Laplacian matrix. 

\paragraph{Dynamic Link Prediction methods.} 
For dynamic link prediction, many methods are based only on \textit{node} representations and use MLPs or cosine similarity to predict the existence of a link \cite{sankar2018dysat:networks, Seo2016StructuredNetworks,rossi2020tgn:graphs}. Recent approaches complement node representations by incorporating pairwise information. Techniques like co-occurrence neighbors matching \cite{yu2023towardslibrary, wang2021inductivewalks} or cross-attention on historical sub-graphs \cite{wang2021tcl:learning} are employed. However, these methods often overlook the global graph structure by focusing on sampled spatio-temporal substructures. For instance, CAW-N~\cite{wang2021inductivewalks} uses anonymous random walks around a pair of nodes and matches their neighborhoods, while DyGformer~\cite{yu2023towardslibrary} applies transformers to one-hop neighborhoods and calculates co-occurrences. These localized approaches fail to capture the broader graph context. TCL \cite{wang2021tcl:learning} is the closest to \ours, using cross-attention between spatio-temporal representations of node pairs. TCL samples historical sub-graphs using BFS and employs contrastive learning for node representation. However, it still relies on sub-graph sampling, missing the full extent of the global graph information. 
In contrast, \ours leverages the entire graph's spectral properties through the supra-Laplacian, incorporating the global structure directly into the spatio-temporal encoding. This holistic approach allows \ours to provide a richer understanding of dynamic interactions, leading to superior link prediction performance.

\section{The SLATE Method}
\label{sec:model}
\begin{figure}[h]
    \centering
    \input{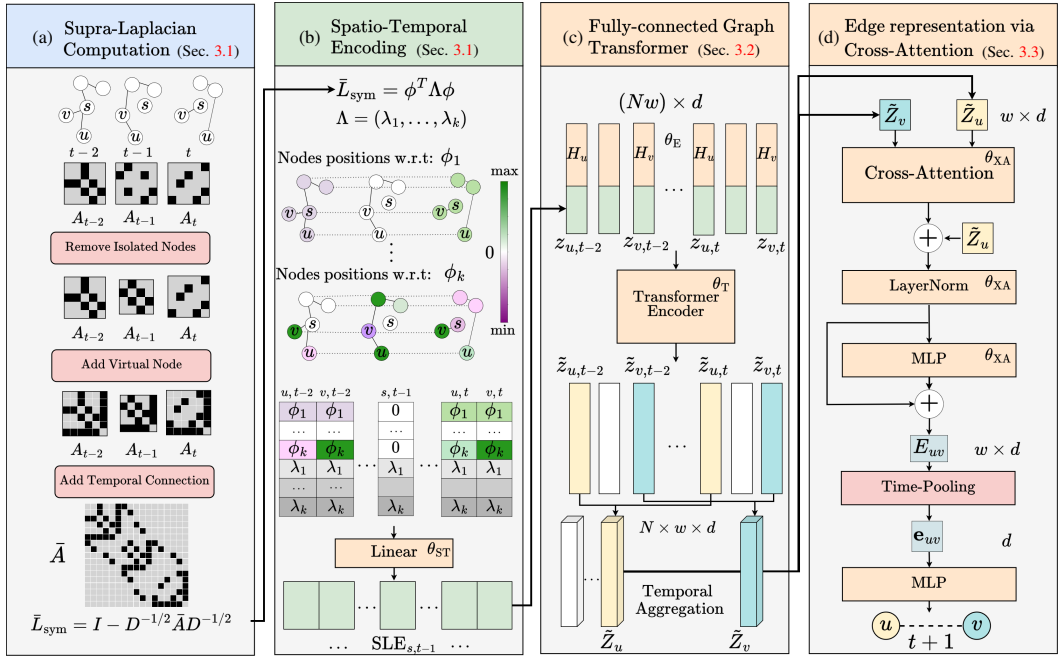}
    \vspace{-1.7em}
   \caption{ The \textbf{\ours model} for link prediction with dynamic graph transformers (DGTs). To recover the lost spatio-temporal structure in DGTs, we adapt the supra-Laplacian matrix computation to DGTs by making the input graph provably connected (a), and use its spectral analysis to introduce a specific encoding for DGTs (b).~(c) Applies a fully connected spatio-temporal transformer between all nodes at multiple time-step. Finally, we design in (d) an edge representations module dedicated to link prediction using cross-attention on multiple temporal representations of the nodes.}
   \label{fig:model}
\end{figure}

In this section, we describe our fully-connected dynamic graph transformer model, \ours, for link prediction.
The core idea in \cref{sec:supralaplacian} is to adapt the supra-Laplacian matrix computations for dynamic graph transformer (DGTs), and to introduce our new spatio-temporal encoding based on its spectral analysis. In \cref{sec:full_attention_transformer}, we detail our full-attention transformer to capture the spatio-temporal dependencies between nodes at different time steps. Finally, we detail our edge representation module for dynamic link prediction in \cref{sec:xa}. \cref{fig:model} illustrates the overall \ours framework.

\textbf{Notations.} Let us consider a DTDG $\mathcal{G}$ as an undirected graph with a fixed number of $N$ nodes across snapshots, represented by the set of adjacency matrices $\mathcal{A} = \{A_1, ..., A_T\}$. Its supra-graph, the multi-layer network $\bar{G}=(\bar{V},\bar{E})$, is associated to a supra-adjacency matrix $\bar{A}$, obtained by stacking $A_i$ diagonally (see \cref{eq:supradj} in~\cref{app:rw_multilayer}). Then, the supra-Laplacian matrix $\bar{L}$ is defined as $\bar{L} = I - \bar{D}^{-1/2}\bar{A}\bar{D}^{-1/2}$, where $I$ is the identity matrix and $\bar{D}$ is the degree matrix of $\bar{G}$. Let $\mathbf{x_u} \in \mathbb{R}^{F}$ be the feature vector associated with the node $u$ (which remains fixed among all snapshots). Finally, let consider the random variable $y \in \{0,1\}$ such that $y=1$ if nodes $u$ and $v$ are connected and $y=0$ otherwise.

\subsection{Supra-Laplacian as Spatio-Temporal Encoding}
\label{sec:supralaplacian}

In this section, we cast Discrete Time Dynamic Graphs (DTDGs) as multi-layer networks, and use the spectral analysis of their supra-graph and generate a powerful spatio-temporal encoding for our fully-connected transformer.

\textbf{DTDG as multi-layer graphs.} If a graph is connected, its spectral analysis provides a rich information of the global graph dynamics, as shown in \cref{fig:supralap}.~The main challenge in casting DTDG as multi-layer graphs relates to its disconnectivity, which induces as many zero eigenvalues as connected components. DTDG have in  practice a high proportion of isolated nodes per snapshot (see \cref{fig:isolated} in experiments), making the spectral analysis on the raw disconnected graph useless. Indeed, it mainly indicates positions relative to isolated nodes, losing valuable information on global dynamics and local spatio-temporal structures. \textcolor{black}{We experimentally validate that it is mandatory to compute the supra-Laplacian matrix on a connected graph to recover a meaningful spatio-temporal structure.} 

\textbf{Supra-Laplacian computation.} To overcome this issue and make the supra-graph connected, we follow three steps: (1) remove isolated nodes in each adjacency matrix, (2) introduce a virtual node in each snapshot to connect clusters, and (3) add a temporal self-connection between a node and its past if it existed in the previous timestep. We avoid temporal dependencies between virtual nodes to prevent artificial connections. These 3 transformation steps make the resulting supra-graph provably connected. This process is illustrated in Figure~\ref{fig:model}\textcolor{black}{a}, and we give the detailed algorithm in \cref{app:supralap}.

\textbf{Spatio-temporal encoding.} With a connected $\mathcal{\bar{G}}$, the second smallest eigenvalue $\lambda_1$ of the supra-Laplacian $\bar{L}$ is guaranteed to be non-negative (see proof in \cref{app:supralap}), and its associated Fiedler vector $\phi_1$ reveals the dynamics of $\mathcal{\bar{G}}$ (\cref{fig:supralap}). In practice, similar to many static GT models \cite{kreuzer2021rethinkingattention,rampasek2022recipetransformer,dwivedi2021graphrepresentations}, we retrieve the first $k$ eigenvectors of the spectrum of $\bar{L}$, with $k$ being a hyper-parameter. The spectrum can be computed in $O(k^2N')$ and have a memory complexity of $O(kN')$ where $N'$ is the size of $\bar{A}$, and we follow the literature to normalize the eigenvectors and resolve sign ambiguities~\cite{kreuzer2021rethinkingattention}. The supra-Laplacian spatio-temporal encoding vector of the node $u$ at time $t$ is:
\begin{equation}
        \label{eq:peconstruction}
        \text{SLE}_{u,t} = \left\{\begin{array}{lr}
        g_{\theta_{ \text{ST} }}(\bar{L}_{u, t} \cdot [\phi_1, \phi_2,...,\phi_k] \oplus \text{diag}(\Lambda))& \text{if $u_{t}$ is not isolated}\\[4pt]
        g_{\theta_{ \text{ST}}}(\mathbf{0}_k   \oplus \text{diag}(\Lambda))& \text{otherwise}\\
        \end{array}\right\}
\end{equation}
where $\oplus$ denotes the concatenation operator. $\bar{L}_{u, t} \cdot [\phi_1, \phi_2,...,\phi_k] = [\phi_1^{u,t},\phi_2^{u,t},...,\phi_k^{u,t}]$ contains the projections of the node $u$ at time $t$ in the eigenspace spanned by the $k$ first eigenvectors of $\bar{L}$, $\text{diag}(\Lambda)$ contains the eigenvalues of $\bar{L}$ (which are the same for all nodes) and $g_{\theta_{\text{ST}}}$ is a linear layer allowing to finely adapt the supra-graph spectrum features to the underlying  link prediction task. Note that because we did not include isolated nodes in the computation of the supra-Laplacian, we replace the eigenvector projections by a null vector $\mathbf{0}^k$ for these nodes. All the steps involved in constructing our spatio-temporal encoding are illustrated in Figure~\ref{fig:model}\textcolor{black}{b}.

\subsection{Fully-connected spatio-temporal transformer}
\label{sec:full_attention_transformer}

In this section, we describe the architecture of our fully-connected spatio-temporal transformer, $f_{\theta_{T}}$, to construct node representations that captures long-range dependencies between the nodes at each time step. We illustrate our fully-connected GT in Figure~\ref{fig:model}\textcolor{black}{c}. We employ a single transformer block, such that our architecture remains lightweight. ~This is in line with recent findings showing that a single encoder layer with multi-head attention is sufficient for high performance, even for dynamic graphs \cite{wu2023simplifyingrepresentations}.

The input representation of the node $u_{t}$ is the concatenation of the node embeddings (which remains the same for each snapshot) and our supra-Laplacian spatio-temporal encoding:

\begin{equation}
\label{eq:tokenconstruction}
    \mathbf{z}_{u,t} =  g_{\theta_{\text{E}}}(\mathbf{x_u}) \oplus \text{SLE}_{u,t}
\end{equation}

where $g_{\theta_{\text{E}}}$ is a linear projection layer and $\oplus$ denotes the concatenation operator. Then we stack all the representations of \textit{each nodes at each time step} within a time window of size $w$ to obtain the input sequence, $Z \in \mathbb{R}^{(Nw) \times d}$, of the GT.

The fully-connected spatio-temporal transformer, $f_{\theta_{T}}$, then produces a unique representation $\tilde{Z} \in \mathbb{R}^{(Nw) \times d}$ for each node at each time-step :
\begin{equation}
    \tilde{Z} = f_{\theta_{T}}(Z).
\end{equation}
Surprisingly, considering all temporal snapshots did not yield better results in our experiments (see \cref{fig:tw} in \cref{sec:model_analysis}).

Unlike previous DGT methods that sample substructures around each nodes \cite{liu2021anomalytransformer, yang2022time-awareevolution, wang2021tcl:learning}, \ours leverages the full structure of the DTDG within the time window. This approach ensures that no nodes are arbitrarily discarded in the representation learning process, as we use the same information source $Z$ for all nodes.

\subsection{Edge Representation with Cross-Attention}
\label{sec:xa}

In this section, we present our innovative edge representation module \link. It is designed for efficient dynamic link prediction and leverage the node representations learned by our fully-connected spatio-temporal GT. We illustrated our module in Figure~\ref{fig:model}\textcolor{black}{d}. This module is composed of a cross-attention model, $f_{\theta_{\text{XA}}}$, that captures pairwise information between the historical representation of two targeted nodes followed by a classifier to determine the presence of a link.

For a link prediction at time $t+1$ on a given node pair $(u, v)$, we aggregate all temporal representations of $u$ and $v$ resulting in two sequences $\tilde{Z}_{u,t}= [\tilde{\mathbf{z}}_{u, t\minus w}, \ldots, \tilde{\mathbf{z}}_{u, t}]$ and $\tilde{Z}_{v,t}= [\tilde{\mathbf{z}}_{v, t\minus w}, \ldots, \tilde{\mathbf{z}}_{v, t}]$.
We use these multiple embeddings to build a pairwise representation that captures dynamic relationships over time. Then, the cross-attention module $f_{\theta_{\text{XA}}}$ produces a pairwise representation of the sequence $E_{u,v} \in \mathbb{R}^{w\times d}$ :
\begin{equation}
    E_{u, v} = f_{\theta_{\text{XA}}}( \tilde{Z}_{u,t},  \tilde{Z}_{v,t}).
\end{equation}

We obtain the final edge representation $\mathbf{e_{u,v}}\in \mathbb{R}^d$ by applying an average time-pooling operator and we compute the probability that the nodes $u_{t+1}$ and $v_{t+1}$ are connected with:
\begin{equation}
    p(y=1|~\mathbf{e_{u, v}}) = \sigma(\text{MLP}(\mathbf{e_{u,v}})).
\end{equation}
\ours differs from methods that enrich node and edge representations with pairwise information by sampling substructures around each node \cite{wang2021tcl:learning,yu2023towardslibrary,wang2021inductivewalks,hu2023recurrentstates}. Instead, we first compute node representations based on the same dynamic graph information contained in $Z$. Then, we capture fine-grained dynamics specific to each link $(u,v)$ through a cross-attention mechanism.

Our training resort to the standard Binary Cross-Entropy loss function. In practice, for a node $u$, we sample a negative pair $v_{\text{neg}}$ and a positive pair $v_{\text{pos}}$: 
\begin{equation}
\label{eq:loss}
    \mathcal{L_\theta} = \text{BCE}(p(y=1| \mathbf{e_{u,v_{\text{pos}}}})) + \text{BCE}(p(y=0| \mathbf{e_{u,v_{\text{neg}}}})) .
\end{equation}
In this context, $\theta =\{\theta_\text{XA},\theta_\text{T},\theta_\text{ST},\theta_\text{E}\}$ represents all the parameters within the edge representation module $\theta_\text{XA}$, the fully-connected  transformer $\theta_T$,  the spatio-temporal linear layer $\theta_{ST}$ and the node embedding parameters $\theta_E$ as illustrated in \cref{fig:model}.

\subsection{SLATE Scalability} \textcolor{black}{The theoretical complexity of attention computation is $O(N^2)$ per snapshot, scaling to $O((NT)^2)$ when considering all $T$ snapshots. However, as shown in our experiments (\cref{fig:tw}) and consistent with recent works \cite{karmim2024temporalreceptivefielddynamic}, a large temporal context is often unnecessary. By using a time window $w$ with $w \ll T$ (similar to other DGT architectures \cite{liu2021anomalytransformer,yang2022time-awareevolution}), we reduce complexity to $O((Nw)^2)$. For predictions at time $t+1$, we focus only on snapshots from $G_{t-w}$ to $G_t$. Ablation studies confirm that smaller time windows deliver excellent results across various real-world datasets. We further leverage FLASH Attention \cite{dao2022flashattention:io-awareness} to optimize memory usage and computation. Additionally, we incorporate Performer \cite{choromanski2022rethinkingattentionperformers}, which approximates the softmax computation of the attention matrix, reducing the complexity to $O(Nw)$. This enables us to scale efficiently to larger graphs, as shown in \cref{tab:efficiency}, while maintaining high performance (see \cref{tab:performer_vs_transformer}) with manageable computational resources.}

\section{Experiments}
\label{sec:experiments}

We conduct extensive
~experiments to validate \ours for link prediction on discrete dynamic graphs, including state-of-the-art comparisons in \cref{sec:sota_compa}. In  \cref{sec:model_analysis}, we highlight the benefits of our two main contributions, the importance of connecting our supra-graph, and the ability of \ours to scale to larger datasets with reasonable time and memory consumption compared to MP-GNNs. 

\textbf{Implementation details.} We use one transformer Encoder Layer \cite{vaswani2017attentionneed}. For larger datasets, we employ Flash Attention \cite{dao2022flashattention:io-awareness} for improved time and memory efficiency. Further details regarding model parameters and their selections are provided in \cref{tab:searchparam}. We fix the token dimension at $d=128$ and the time window at $w=3$ for all our experiments. We use an SGD optimizer for all of our experiments. Further details on hyper-parameters search, including the number of eigenvectors for our spatio-temporal encoding, are in \cref{app:param}.

\subsection{Comparison to state-of-the-art}
\label{sec:sota_compa}

Since both the continuous and discrete communities evaluate on similar data, we compare \ours to state-of-the-art DTDG (\cref{tab:dtdg_main_auc}) and CTDG (\cref{tab:ctdg_main_auc}) models. Best results are in bold, second best are underlined. More detailed results and analyses are presented in \cref{app:additionnalexp}.

\textbf{Baselines and evaluation protocol.} To compare the benefits of fully connected spatio-temporal attention with a standard approach using transformers, we designed the ROLAND-GT model based on the ROLAND framework \cite{you2022roland:graphs}. This model follows the stacked-GNN approach \cite{skarding2021foundationssurvey}, equipped with the encoder $f_{\theta_T}$ described in \cref{sec:model} including static Laplacian positional encoding \cite{dwivedi2020agraphs}, and a LSTM \cite{hochreiter1997longmemory} updating the node embeddings.

We adhere to the standardized evaluation protocols for continuous models~\cite{yu2023towardslibrary} and discrete models~\cite{yang2021discrete-timespace}. Our evaluation follows these protocols, including metrics, data splitting, and the datasets provided. Results in \cref{tab:dtdg_main_auc} and \cref{tab:ctdg_main_auc} are from the original papers, except those marked with $^\dagger$. We report the average results and standard deviations from five runs to assess robustness. Additional results, including hard negative sampling evaluation, are in \cref{app:hardnss}.

\textbf{Datasets.} In \cref{tab:datasets}  \cref{app:dts}, we provide detailed statistics for the datasets used in our experiments.  An in-depth description of the datasets is given in \cref{app:dts}. We evaluate on DTDGs datasets provided by \cite{yu2023towardslibrary} and \cite{yang2021discrete-timespace}, we add a synthetic dataset SBM based on stochastic block model \cite{lee2019aclustering}, to evaluate on denser DTDG.

\begin{table*}[!ht]
    \setlength\tabcolsep{10pt}
    \setlength\extrarowheight{0.5pt}
    \caption{Comparison to DTDG models on discrete data. ROC-AUC}
    \small
    \label{tab:dtdg_main_auc} 
    \centering
    \resizebox{\linewidth}{!}{
    \begin{tabular}{  c | c | c | c | c | c | c }
        \toprule
         Method & HepPh  & AS733 & Enron & Colab & SBM$^\dagger$ & \textbf{Avg.} \\

        \midrule
          GCN$^\dagger$ \cite{kipf2016semi-supervisednetworks} & 74.52 \small{± 0.80}  & 96.65 \small{± 0.05} & 91.31 \small{± 0.45} & 88.28 \small{± 0.58} & \underline{95.96} \small{± 0.32} & 89.34 \small{± 0.44}\\
          GIN$^\dagger$ \cite{xu2018hownetworks} & 71.47 \small{± 0.56}  & 93.53 \small{± 0.55} & 91.16 \small{± 1.17} & 85.38 \small{± 0.61} & 88.86 \small{± 0.46} & 86.08 \small{± 0.67 } \\
          EvolveGCN \cite{pareja2019evolvegcn:graphs}  & 76.82 \small{± 1.46}  & 92.47 \small{± 0.04} & 90.12 \small{± 0.69} & 83.88 \small{± 0.53} & 94.21 \small{± 0.66} & 87.50 \small{± 0.68} \\
          GRUGCN \cite{Seo2016StructuredNetworks} & 82.86 \small{± 0.53}  & 94.96 \small{± 0.35} & 92.47 \small{± 0.36} & 84.60 \small{± 0.92} & 92.55 \small{± 0.41} & 89.48 \small{± 0.51} \\
          DySat \cite{sankar2018dysat:networks} & 81.02 \small{± 0.25}  & 95.06 \small{± 0.21} & 93.06 \small{± 0.97} & 87.25 \small{± 1.70} & 91.92 \small{± 0.39} & 89.67 \small{± 0.70} \\
          VGRNN \cite{hajiramezanali2019variationalnetworks} & 77.65 \small{± 0.99}  & 95.17 \small{± 0.62} & 93.10 \small{± 0.57} &85.95 \small{± 0.49} & 93.88 \small{± 0.07} & 89.15 \small{± 0.55} \\
          HTGN \cite{yang2021discrete-timespace}& 91.13 \small{± 0.14}  & \textbf{98.75} \small{± 0.03} & \underline{94.17} \small{± 0.17} & \underline{89.26} \small{± 0.17} & 94.80 \small{± 0.23} & \underline{93.62} \small{± 0.15} \\
          ROLAND-GT$^\dagger$ \cite{you2022roland:graphs} & 81.40 \small{± 0.45}  & 94.75 \small{± 0.87} & 90.20 \small{± 1.12} & 82.95 \small{± 0.45} & 94.88 \small{± 0.31} & 88.83 \small{± 0.64} \\
          \midrule
          \rowcolor{slatecolor}
    \textbf{\ours} & \textbf{93.21} \small{± 0.37} & \underline{97.46} \small{± 0.45} & \textbf{96.39} \small{± 0.17} &\textbf{90.84} \small{± 0.41} & \textbf{97.69} \small{± 0.21} & \textbf{95.12} \small{± 0.32} \\
    \bottomrule
    \end{tabular}
    }
\end{table*}
\textbf{Comparison to discrete models, on DTDG. \cref{tab:dtdg_main_auc}}
 We showcases the performance of \ours against various discrete models on DTDG datasets, highlighting its superior performance across multiple metrics and datasets. \ours outperforms all state of the art models on the HepPh, Enron, and Colab datasets, demonstrating superior dynamic link prediction capabilities. Notably, it surpasses HTGN by +2.1 points in AUC on HepPh and +1.1 points in AP on Enron. Moreover, \ours shows a remarkable improvement of +7.6 points in AUC over EvolveGCN on Colab. It also performs competitively on the AS733 dataset, with scores that are closely second to HTGN, demonstrating its robustness across different types of dynamic graphs.
What also emerges and validates our method from this comparison is the average gain of +6.29 points by our fully connected spatio-temporal attention model over the separate spatial attention model and temporal model approach, as used in ROLAND-GT. We also demonstrate significant gains against sparse attention models like DySat, with an increase of +6.45.
This study, conducted on the protocol from \cite{yang2021discrete-timespace}, emphasizes \ours
~capability in handling discrete-time dynamic graph data, offering significant improvements over existing models.

\begin{table*}[!ht]
    \small
    \setlength\tabcolsep{7pt}
    \setlength\extrarowheight{0.5pt}
    \caption{Comparison to CTDG models on discrete data using \cite{yu2023towardslibrary} protocol (AUC).}
    \label{tab:ctdg_main_auc}
    \centering
    \resizebox{\linewidth}{!}{
    \begin{tabular}{  c | c | c | c | c | c | c | c}
        \toprule
          Method & CanParl  & USLegis & Flights & Trade & UNVote & Contact & \textbf{Avg.}\\

        \midrule
           JODIE \cite{kumar2019jodie:networks} &  78.21 \text{± 0.23}  & 82.85 \text{± 1.07} &96.21 \text{± 1.42} &69.62 \text{± 0.44} &68.53 \text{± 0.95} &96.66 \text{± 0.89} & 82.01 ± 0.83\\
          DyREP \cite{trivedi2018dyrep:graphs}  & 73.35 \text{± 3.67}  & 82.28\text{ ± 0.32} & 95.95 \text{± 0.62} & 67.44 \text{± 0.83} & 67.18 \text{± 1.04} & 96.48 \text{± 0.14} & 80.45 ± 1.10\\
          TGAT \cite{xu2020tgat:graphs}  & 75.69 \text{± 0.78} &  75.84 \text{± 1.99} & 94.13 \text{± 0.17} &64.01 \text{± 0.12} & 52.83 \text{± 1.12}& 96.95 \text{± 0.08} & 76.58 ± 0.71\\
          TGN \cite{rossi2020tgn:graphs} & 76.99 \text{± 1.80} & \underline{83.34} \text{± 0.43} &98.22 \text{± 0.13} &69.10 \text{± 1.67} & \underline{69.71} \text{± 2.65}& 97.54 \text{± 0.35} & 82.48 ± 1.17\\
          CAWN \cite{wang2021inductivewalks} & 75.70\text{ ± 3.27} & 77.16 \text{± 0.39} &{98.45 \text{± 0.01}} & 68.54 \text{± 0.18}&53.09 \text{± 0.22} & 89.99 \text{± 0.34} & 77.16 ± 0.74\\
          EdgeBank \cite{poursafaei2022towardsprediction} & 64.14 \text{± 0.00}  & 62.57 \text{± 0.00} &90.23\text{ ± 0.00} & 66.75 \text{± 0.00} & 62.97 \text{± 0.00} & 94.34 \text{± 0.00} & 73.50 ± 0.00\\
          TCL \cite{wang2021tcl:learning} & 72.46 \text{± 3.23} & 76.27 \text{± 0.63} &91.21 \text{± 0.02} &64.72 \text{± 0.05} &51.88 \text{± 0.36} & 94.15 \text{± 0.09} & 75.11 ± 0.73\\
          GraphMixer \cite{cong2023graphmixer:networks} & 83.17 \text{± 0.53} & 76.96 \text{± 0.79}  & 91.13 \text{± 0.01}& 65.52 \text{± 0.51}&52.46 \text{± 0.27} & 93.94 \text{± 0.02} & 77.20 ± 0.36\\
          DyGformer \cite{yu2023towardslibrary} & \textbf{97.76 }\text{± 0.41} &  77.90 \text{± 0.58} & \underline{98.93} \text{± 0.01} &\underline{70.20} \text{± 1.44} & 57.12 \text{± 0.62} & \textbf{98.53} \text{± 0.01} & \underline{83.41} ± 0.51\\
          \midrule
          \rowcolor{slatecolor} \textbf{\ours}   &\underline{92.37} \text{± 0.51}& \textbf{95.80} \text{± 0.11} & \textbf{99.07} \text{± 0.41} &\textbf{96.73} \text{± 0.29} &\textbf{99.94} \text{± 0.05} & \underline{98.12} \text{± 0.37} & \textbf{96.88} ± 0.26 \\
         \bottomrule
    \end{tabular}
    }
\end{table*}

\textbf{Comparison to continuous models, on DTDG. \cref{tab:ctdg_main_auc}}
In dynamic link prediction, \ours outperforms models focused on node (TGN, DyRep, TGAT), edge (CAWN), and combined node-pairwise information (DyGFormer,TCL). Notably, it surpasses TCL by over 21 points in average, showcasing the benefits of our temporal cross attention strategies. \ours's advantage stems from its global attention mechanism, unlike the sparse attention used by TGAT, TGN, and TCL. By employing fully-connected spatio-temporal attention, \ours directly leverages temporal dimensions through its \link module. This strategic approach allows \ours to excel, as demonstrated by its consistent top performance and further evidenced in Appendix with hard negative sampling results (see \cref{tab:ctdg_full_ap} and \cref{tab:ctdg_full_auc} in \cref{app:additionnalexp}). We demonstrate average results that are superior by 13~points compared to the most recent model on DTDG, DyGFormer~\cite{yu2023towardslibrary}.

\subsection{Model Analysis}
\label{sec:model_analysis}
\textbf{Impact of different \ours component.} Table \ref{tab:impact_components} presents the AUC results of different configurations of \ours on four datasets. This evaluation demonstrates the impact of our proposed spatio-temporal encoding and the \link module on dynamic link prediction performance.
\begin{table}[!ht]
\small
    \centering
    \setlength\tabcolsep{7pt}
    \setlength\extrarowheight{0.5pt}
    \resizebox{\linewidth}{!}{
    \begin{tabular}{cc|cccc}
    \toprule
         Encoding & \link Module & Enron & CanParl & USLegis & UNtrade\\
         \midrule
         LapPE\cite{dwivedi2020agraphs} + sinus-based \cite{vaswani2017attentionneed} & \xmark & 89.18 ± 0.33 & 82.98 ± 0.71 & 85.22 ± 0.24 & 90.24 ± 1.05 \\
          \ours  & \xmark & {90.57}  ± 0.27 & {89.45}  ± 0.38 & {93.30} ± 0.29 & {94.01} ± 0.73 \\
         LapPE \cite{dwivedi2020agraphs} + sinus-based \cite{vaswani2017attentionneed} & \cmark & 90.75 ± 0.08 &  90.23 ± 0.41  & 87.50 ± 0.50 & 90.56 ± 0.69 \\
         \midrule
         \rowcolor{slatecolor} \textbf{\ours}  & \cmark & \textbf{96.39} ± 0.18 & \textbf{92.37} ± 0.51 & \textbf{95.80} ± 0.11 & \textbf{96.73} ± 0.29 \\
         \bottomrule
    \end{tabular}}
    \caption{Validation of different \ours component. Results in AUC over 4 datasets. }
    \label{tab:impact_components}
\end{table}

First, we show the naive spatio-temporal encoding approach using the first $k$ Laplacian eigenvectors associated with the $k$ lowest values \cite{dwivedi2020agraphs} (\cref{app:lappe}), combined with sinusoidal unparametrized temporal encoding  \cite{vaswani2017attentionneed} (\cref{app:timepe}), without the \link module. The Laplacian is computed sequentially on the $w$ snapshots, then concatenated with the temporal encoding indicating the position of the snapshot, with $k=12$ for both \ours and the naive encoding. The AUC scores across all datasets are significantly lower, highlighting the limitations of this naive encoding method in capturing complex spatio-temporal dependencies.

Replacing the baseline encoding with our proposed \ours encoding, still without the \link module, results in significant improvements: +6.47 points on CanParl, +8.08 points on USLegis, and +3.77 points on UNtrade. These improvements demonstrate the effectiveness of our spatio-temporal encoding. Adding the \link module to the naive encoding baseline yields further improvements: +7.25 points on CanParl and +1.57 points on Enron. However, it still falls short compared to the enhancements provided by the \ours encoding.

Finally, the complete model, \ours with the \link module, achieves the highest AUC scores across all datasets: +9.39 points on CanParl and +10.58 points on USLegis. These substantial gains confirm that integrating our unified spatio-temporal encoding and the \link module effectively captures intricate dynamics between nodes over time, resulting in superior performance.

\begin{figure}[!ht]
    \centering
    \begin{minipage}{0.45\textwidth} 
        \centering
        \small
        \begin{tabular}{l|cc}
        \toprule
            Dataset & SLATE w/o trsf &  \cellcolor{slatecolor} \textbf{\ours} \\
            \midrule
            Colab & 85.03 ± 0.72 & \cellcolor{slatecolor}\textbf{90.84} ± 0.41 \\
            USLegis & 63.35 ± 1.24 & \cellcolor{slatecolor}\textbf{95.80} ± 0.11 \\
            UNVote & 78.30 ± 2.05 & \cellcolor{slatecolor}\textbf{99.94} ± 0.05 \\
            AS733 & 81.50 ± 1.35 & \cellcolor{slatecolor}\textbf{97.46} ± 0.45 \\
            \bottomrule
        \end{tabular}
        \captionof{table}{Importance of connectivity transformations steps to connect the supra-adjacency matrix. AUC performance in dynamic link prediction.}
        \label{tab:process_adj}
    \end{minipage}%
    \hspace{0.05\textwidth} 
    \begin{minipage}{0.45\textwidth} 
        \centering
        \includegraphics[width=\textwidth]{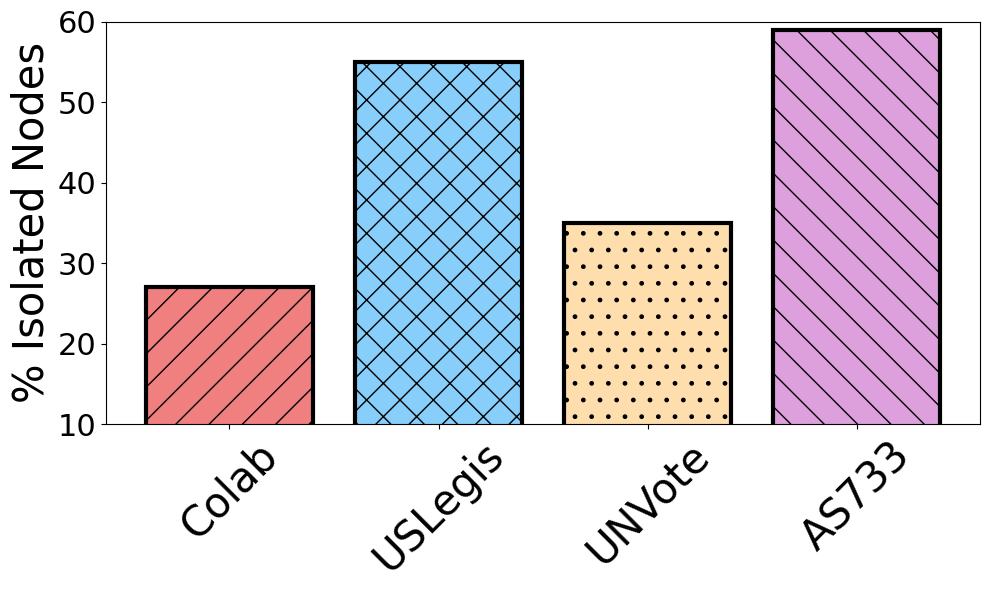}
        \vspace{-1.5em}
        \captionof{figure}{Average percentage of isolated nodes per snapshot on real world dynamic graphs data.}
        \label{fig:isolated}
    \end{minipage}
\end{figure}

\textbf{Critical role of supra-adjacency transformation.} Here, we demonstrate the importance of the transformation steps of the supra-adjacency matrix, as detailed in \cref{sec:supralaplacian}, by removing isolated nodes, adding virtual nodes, and incorporating temporal connections (\cref{fig:model}). Table \ref{tab:process_adj} presents the performance of \ours with and without transformation (trsf) on four datasets. Without these critical transformations, there is a systematic drop in performance, particularly pronounced in datasets with a high number of isolated nodes, as shown in \cref{fig:isolated} (27\% in Colab, 53\% in USLegis, 35\% in UNVote, and 59\% in AS733). These results clearly highlight the significant improvements brought by our proposed transformations. \textcolor{black}{More detailed experiments regarding each transformation, particularly on the importance of removing isolated nodes and adding a virtual node, are presented in \cref{tab:isolated_vs_slate,tab:with_wo_vn}.}

\begin{figure}[!htb]
    \centering 
    \begin{minipage}{0.45\textwidth} 
        \small
        \centering
        \setlength{\tabcolsep}{3pt}
        \begin{tabular}{l|llc}
        \toprule
            Models & Mem. & t / ep.  & Nb params.  \\
        \midrule
            EvolveGCN & 46Go & 1828s & 1.8 M \\
            DySAT & 42Go & 1077s & 1.8 M \\
            VGRNN & 21Go & 931s & \textbf{0.4 M} \\
            ROLAND-GT \small{w/o Flash}& OOM & - & 1.9 M \\
            ROLAND-GT & 44Go & 1152s & 1.9 M \\
            \midrule
            SLATE \small{w/o Flash} & OOM & - & 2.1 M \\
            SLATE & 48Go & 1354s & 2.1 M  \\
            \rowcolor{slatecolor} SLATE-Performer & \textbf{17Go} & \textbf{697s} & 2.1 M \\
        \bottomrule
        \end{tabular}
        \captionof{table}{An analysis of model efficiency comparing the memory usage (Mem.), training time per epoch (t/ep.) and the number of parameters (Nb params) on Flights dataset}
        \label{tab:efficiency}
    \end{minipage}\hfill
    \begin{minipage}{0.48\textwidth}
        \centering
        \includegraphics[width=\textwidth]{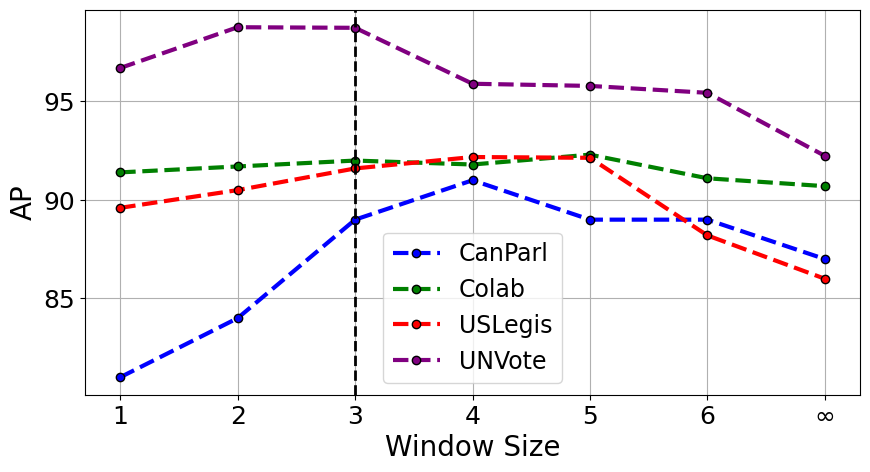}
        \caption{Model performance based on the window size, $w = \infty$ corresponds to considering all snapshots. Results in average precision (AP). }
        \label{fig:tw}
    \end{minipage}
    \vspace{-1.5em}
\end{figure}

\paragraph{Impact of the time-window size.} We demonstrate in \cref{fig:tw} the impact of the time window size on the performance of the \ours model. A window size of 1 is equivalent to applying a global attention transformer to the latest snapshot before prediction, and an infinite window size is equivalent to considering all the snapshots for global attention. This figure highlights the importance of temporal context for accurate predictions within dynamic graphs. We observe that, in most cases, too much temporal context can introduce noise into the predictions. The USLegis, UNVote and CanParl datasets are political graphs spanning decades (72 years for UNVote), making it unnecessary to look too far back. For all of our main results in \cref{tab:ctdg_main_auc} and \cref{tab:dtdg_main_auc} we fix for simplicity $w=3$. However, our ablations have identified $w=4$ as an optimal balance, capturing sufficient temporal context without introducing noise into the transformer encoder and ensuring scalability for our model. Therefore, \ours performances could further be improved by more systematic cross-validation of its hyper-parameters, \eg $w$.

\vspace{-0.8em}
\paragraph{Model efficiency.}
The classic attention mechanism, with a complexity of $O(N^2)$, can be memory-consuming when applied across all nodes at different time steps. However, using Flash-Attention~\cite{dao2022flashattention:io-awareness} and a light transformer architecture with just one encoder layer, we successfully scaled to the Flights dataset, containing 13,000 nodes and a window size of $w=3$. \textcolor{black}{By using the Performer encoder \cite{choromanski2022rethinkingattentionperformers}, which approximates attention computation with linear complexity, memory usage is reduced to 17GB}. Our analysis shows that our model empirically matches the memory consumption of various DTDG architectures while maintaining comparable computation times (\cref{tab:efficiency}). Furthermore, it is not over-parameterized relative to existing methods. We trained on an NVIDIA-Quadro RTX A6000 with 49 GB of total memory.
\subsection{Qualitative results}
\begin{figure}[!ht]
        \centering
        \includegraphics[width=0.75\textwidth]{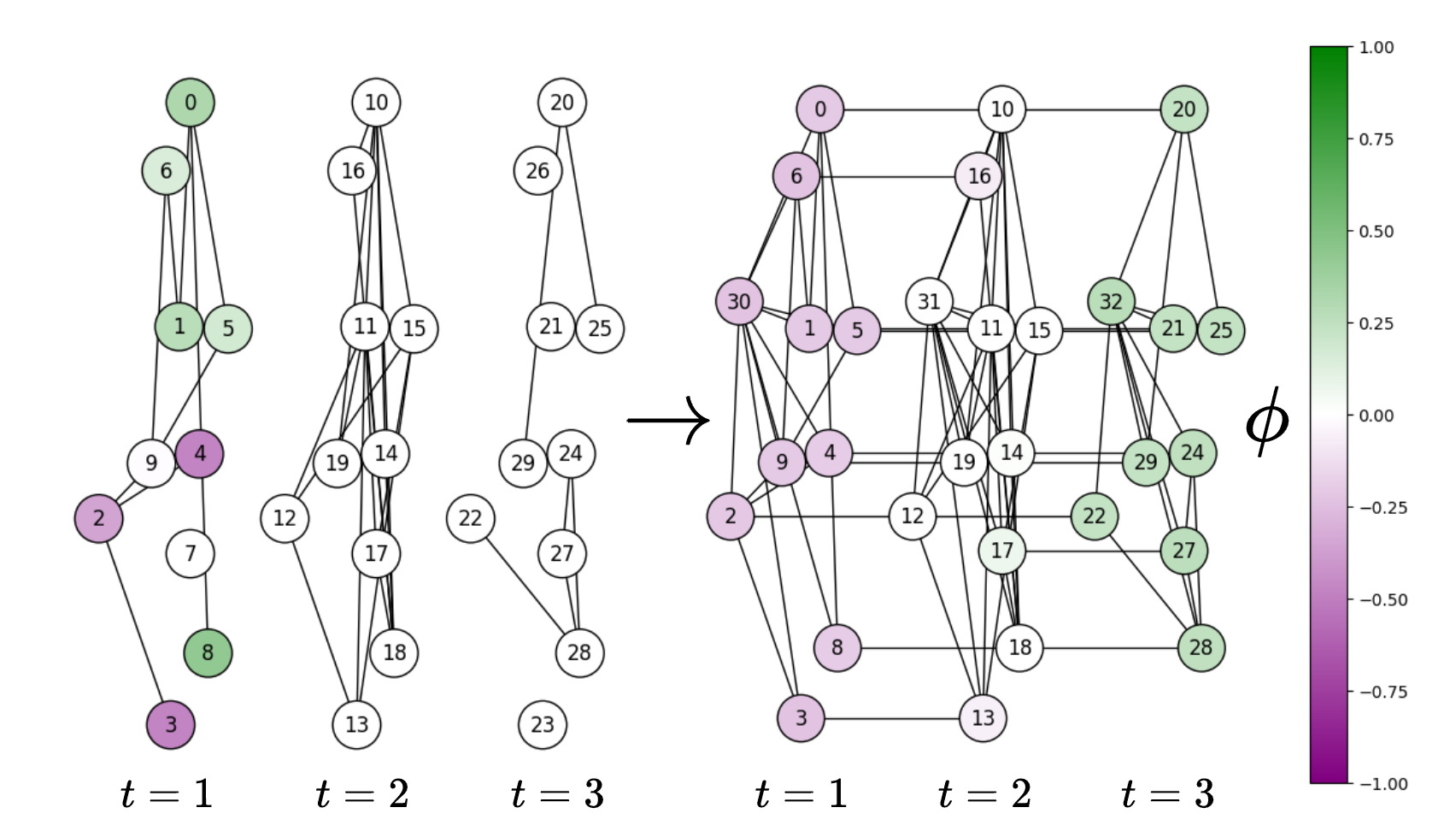}
        \caption{Projection of the eigenvector associated with the first non-zero eigenvalue on a toy DTDG before and after transformation. On the \textit{left}, the DTDG is unprocessed, showing only spatial projections due to the lack of temporal connections. On the \textit{right}, after applying the \ours transformation, the graph captures rich spatio-temporal properties, allowing for a more informative spatio-temporal encoding}
        \label{fig:spectrum_toy}
    \end{figure}

\textcolor{black}{We present qualitative results in \cref{fig:spectrum_toy} comparing the graph and its spectrum before and after applying the proposed transformation in \ours. The projection is made on the eigenvector associated with the first non-zero eigenvalue. Before transformation, the DTDG contains isolated nodes (7, 23 and 26) and two distinct clusters in the snapshot at $t=3$. In this case, the projection is purely spatial, as there are no temporal connections, and some projections also occur on isolated nodes due to the presence of distinct connected components. After the proposed transformation into a connected multi-layer graph, the projection captures richer spatio-temporal properties of the dynamic graph. By connecting the clusters with a virtual node and adding temporal edges, our approach removes the influence of isolated nodes and enables the construction of an informative spatio-temporal encoding that better reflects the dynamic nature of the graph.}

\section{Conclusion}
\label{sec:conclu}

We have presented the \ours method, an innovative spatio-temporal encoding for transformers on dynamic graphs, based on supra-Laplacian analysis. Considering discrete-time dynamic graphs as multi-layer networks, we devise an extremely efficient unified spatio-temporal encoding thanks to the spectral properties of the supra-adjacency matrix. We integrate this encoding into a fully-connected transformer. 
By modeling pairwise relationships in a new edge representation module, we show how it enhances link prediction on dynamic graphs. \ours performs better than previous state-of-the-art approaches on various standard benchmark datasets, setting new state-of-the-art results for discrete link prediction. 

Despite its strong performances, \ours currently operates in a transductive setting and cannot generalize to unseen nodes. We aim to explore combinations with MP-GNNs to leverage the strengths of local feature aggregation and global contextual information. On the other hand, \ours scales reasonably well to graphs up to a certain size but, as is often the case with transformers, future work is required to scale to very large graphs. 

\section{Acknowledgement}
We acknowledge the financial support provided by PEPR Sharp (ANR-23-PEIA-0008, ANR, FRANCE 2030). We would also like to thank the LITIS laboratory in Rouen and especially Leshanshui Yang, who helped us better position our method. We also thank Elias Ramzi for his feedback on the paper and assistance in writing the abstract.
\bibliographystyle{plain}
\bibliography{references}

\begin{thebibliography}{10}

\bibitem{alon2020onimplications}
Uri Alon and Eran Yahav.
\newblock On the bottleneck of graph neural networks and its practical implications.
\newblock In {\em 9th International Conference on Learning Representations, {ICLR} 2021, Virtual Event, Austria, May 3-7, 2021}, 2021.

\bibitem{banerjee2022spatial-temporalforecasting}
Soumyanil Banerjee, Ming Dong, and Weisong Shi.
\newblock {Spatial-Temporal Synchronous Graph Transformer network (STSGT) for COVID-19 forecasting}.
\newblock {\em Smart Health}, 26, October 2022.

\bibitem{beaini2020directionalnetworks}
Dominique Beaini, Saro Passaro, Vincent L{\'e}tourneau, Will Hamilton, Gabriele Corso, and Pietro Li{\`o}.
\newblock Directional graph networks.
\newblock In {\em International Conference on Machine Learning}, pages 748--758. PMLR, 2021.

\bibitem{chen2020cansubstructures}
Zhengdao Chen, Lei Chen, Soledad Villar, and Joan Bruna.
\newblock Can graph neural networks count substructures?
\newblock In {\em Advances in neural information processing systems}, volume~33, pages 10383--10395, 2020.

\bibitem{choromanski2022rethinkingattentionperformers}
Krzysztof Choromanski, Valerii Likhosherstov, David Dohan, Xingyou Song, Andreea Gane, Tamas Sarlos, Peter Hawkins, Jared Davis, Afroz Mohiuddin, Lukasz Kaiser, David Belanger, Lucy Colwell, and Adrian Weller.
\newblock Rethinking attention with performers, 2022.

\bibitem{chutransmot:tracking}
Peng Chu, Jiang Wang, Quanzeng You, Haibin Ling, and Zicheng Liu.
\newblock Transmot: Spatial-temporal graph transformer for multiple object tracking.
\newblock In {\em Proceedings of the IEEE/CVF Winter Conference on applications of computer vision}, pages 4870--4880, 2023.

\bibitem{cong2023graphmixer:networks}
Weilin Cong, Si~Zhang~Meta, Jian Kang, Baichuan Yuan, Hao Wu, Xin Zhou~Meta, Hanghang Tong, and Mehrdad Mahdavi.
\newblock {GraphMixer: Do We Really Need Complicated Model Architectures For Temporal Networks?}
\newblock {\em arXiv preprint arXiv:2302.11636}, 2023.

\bibitem{cozzomultilayerproperties}
Emanuele Cozzo, Guilherme~Ferraz de~Arruda, Francisco~A Rodrigues, and Yamir Moreno.
\newblock Multilayer networks: metrics and spectral properties.
\newblock {\em Interconnected networks}, pages 17--35, 2016.

\bibitem{dao2022flashattention:io-awareness}
Tri Dao, Dan Fu, Stefano Ermon, Atri Rudra, and Christopher R{\'e}.
\newblock Flashattention: Fast and memory-efficient exact attention with io-awareness.
\newblock In {\em Advances in Neural Information Processing Systems}, volume~35, pages 16344--16359, 2022.

\bibitem{dong2014clusteringmanifolds}
Xiaowen Dong, Pascal Frossard, Pierre Vandergheynst, and Nikolai Nefedov.
\newblock {Clustering on multi-layer graphs via subspace analysis on Grassmann manifolds}.
\newblock {\em IEEE Transactions on Signal Processing}, 62(4):905--918, 2 2014.

\bibitem{dwivedi2020agraphs}
Vijay~Prakash Dwivedi and Xavier Bresson.
\newblock {A Generalization of Transformer Networks to Graphs}.
\newblock {\em arXiv preprint arXiv:2012.09699}, 2020.

\bibitem{dwivedi2021graphrepresentations}
Vijay~Prakash Dwivedi, Anh~Tuan Luu, Thomas Laurent, Yoshua Bengio, and Xavier Bresson.
\newblock Graph neural networks with learnable structural and positional representations.
\newblock In {\em The Tenth International Conference on Learning Representations, {ICLR} 2022, Virtual Event, April 25-29, 2022}. OpenReview.net, 2022.

\bibitem{erddsoni}
Paul Erd{\H{o}}s and Alfr{\'e}d R{\'e}nyi.
\newblock On random graphs {I}.
\newblock {\em Publ. math. debrecen}, 6(290-297):18, 1959.

\bibitem{Fiedler1973}
Miroslav Fiedler.
\newblock Algebraic connectivity of graphs.
\newblock {\em Czechoslovak Mathematical Journal}, 23(2):298--305, 1973.

\bibitem{hajiramezanali2019variationalnetworks}
Ehsan Hajiramezanali, Arman Hasanzadeh, Krishna Narayanan, Nick Duffield, Mingyuan Zhou, and Xiaoning Qian.
\newblock Variational graph recurrent neural networks.
\newblock {\em Advances in neural information processing systems}, 32, 2019.

\bibitem{he2020lightgcn:recommendation}
Xiangnan He, Kuan Deng, Xiang Wang, Yan Li, Yongdong Zhang, and Meng Wang.
\newblock Lightgcn: Simplifying and powering graph convolution network for recommendation.
\newblock In {\em Proceedings of the 43rd International ACM SIGIR conference on research and development in Information Retrieval}, pages 639--648, 2020.

\bibitem{hochreiter1997longmemory}
Sepp Hochreiter and Jürgen Schmidhuber.
\newblock {Long Short-Term Memory}.
\newblock {\em Neural Computation}, 9(8):1735--1780, 11 1997.

\bibitem{hu2023recurrentstates}
Shengxiang Hu, Guobing Zou, Shiyi Lin, Liangrui Wu, Chenyang Zhou, Bofeng Zhang, and Yixin Chen.
\newblock {Recurrent Transformer for Dynamic Graph Representation Learning with Edge Temporal States}.
\newblock {\em arXiv preprint arXiv:2304.10079v1}, 4 2023.

\bibitem{jumper2021highlyalphafold}
John Jumper, Richard Evans, Alexander Pritzel, Tim Green, Michael Figurnov, Olaf Ronneberger, Kathryn Tunyasuvunakool, Russ Bates, Augustin {\v{Z}}{\'{i}}dek, Anna Potapenko, Alex Bridgland, Clemens Meyer, Simon~A.A. Kohl, Andrew~J. Ballard, Andrew Cowie, Bernardino Romera-Paredes, Stanislav Nikolov, Rishub Jain, Jonas Adler, Trevor Back, Stig Petersen, David Reiman, Ellen Clancy, Michal Zielinski, Martin Steinegger, Michalina Pacholska, Tamas Berghammer, Sebastian Bodenstein, David Silver, Oriol Vinyals, Andrew~W. Senior, Koray Kavukcuoglu, Pushmeet Kohli, and Demis Hassabis.
\newblock {Highly accurate protein structure prediction with AlphaFold}.
\newblock {\em Nature 2021 596:7873}, 596(7873):583--589, 7 2021.

\bibitem{kaba2022equivariant}
Oumar Kaba and Siamak Ravanbakhsh.
\newblock Equivariant networks for crystal structures.
\newblock {\em Advances in Neural Information Processing Systems}, 35:4150--4164, 2022.

\bibitem{karmim2024temporalreceptivefielddynamic}
Yannis Karmim, Leshanshui Yang, Raphaël~Fournier S'Niehotta, Clément Chatelain, Sébastien Adam, and Nicolas Thome.
\newblock Temporal receptive field in dynamic graph learning: A comprehensive analysis, 2024.

\bibitem{kim2022purelearners}
Jinwoo Kim, Dat Nguyen, Seonwoo Min, Sungjun Cho, Moontae Lee, Honglak Lee, and Seunghoon Hong.
\newblock Pure transformers are powerful graph learners.
\newblock {\em Advances in Neural Information Processing Systems}, 35:14582--14595, 2022.

\bibitem{kipf2016semi-supervisednetworks}
Thomas~N. Kipf and Max Welling.
\newblock {Semi-Supervised Classification with Graph Convolutional Networks}.
\newblock {\em arXiv preprint arXiv:1609.02907}, 9 2016.

\bibitem{kipf2016variationalauto-encoders}
Thomas~N. Kipf and Max Welling.
\newblock {Variational Graph Auto-Encoders}.
\newblock {\em arXiv preprint arXiv:1611.07308}, 11 2016.

\bibitem{kivela2014multilayer}
Mikko Kivel{\"a}, Alex Arenas, Marc Barthelemy, James~P Gleeson, Yamir Moreno, and Mason~A Porter.
\newblock Multilayer networks.
\newblock {\em Journal of complex networks}, 2(3):203--271, 2014.

\bibitem{kreuzer2021rethinkingattention}
Devin Kreuzer, Dominique Beaini, William~L. Hamilton, Vincent L{\'{e}}tourneau, and Prudencio Tossou.
\newblock {Rethinking Graph Transformers with Spectral Attention}.
\newblock {\em Advances in Neural Information Processing Systems}, 26:21618--21629, 6 2021.

\bibitem{kumar2019jodie:networks}
Srijan Kumar, Xikun Zhang, and Jure Leskovec.
\newblock {JODIE: Predicting dynamic embedding trajectory in temporal interaction networks}.
\newblock {\em Proceedings of the ACM SIGKDD International Conference on Knowledge Discovery and Data Mining}, pages 1269--1278, 8 2019.

\bibitem{kuo2023dynamictransformer}
Ai-Te Kuo, Haiquan Chen, Yu-Hsuan Kuo, and Wei-Shinn Ku.
\newblock {Dynamic Graph Representation Learning for Depression Screening with Transformer}.
\newblock {\em arXiv preprint arXiv:2305.06447v1}, 5 2023.

\bibitem{lee2019aclustering}
Clement Lee and Darren~J. Wilkinson.
\newblock {A review of stochastic block models and extensions for graph clustering}.
\newblock {\em Applied Network Science 2019 4:1}, 4(1):1--50, 12 2019.

\bibitem{li2019predictinggraphs}
Jia Li, Jiao Su, Zhichao Han, Pengyun Wang, Lujia Pan, Hong Cheng, and Jianfeng Zhang.
\newblock {Predicting Path Failure In Time-Evolving Graphs}.
\newblock {\em Proceedings of the ACM SIGKDD International Conference on Knowledge Discovery and Data Mining}, pages 1279--1289, 5 2019.

\bibitem{liu2021anomalytransformer}
Yixin Liu, Shirui Pan, Yu~Guang Wang, Fei Xiong, Liang Wang, Qingfeng Chen, and Vincent~CS Lee.
\newblock {Anomaly Detection in Dynamic Graphs via Transformer}.
\newblock {\em IEEE Transactions on Knowledge and Data Engineering}, 35(12):12081--12094, 2021.

\bibitem{mialon2021graphit:transformers}
Grégoire Mialon, Dexiong Chen, Margot Selosse, and Julien Mairal.
\newblock {GraphiT: Encoding Graph Structure in Transformers}.
\newblock {\em arXiv preprint arXiv:2106.05667v1}, 6 2021.

\bibitem{morris2018weisfeilernetworks}
Christopher Morris, Martin Ritzert, Matthias Fey, William~L Hamilton, Jan~Eric Lenssen, Gaurav Rattan, and Martin Grohe.
\newblock Weisfeiler and leman go neural: Higher-order graph neural networks.
\newblock In {\em Proceedings of the AAAI conference on artificial intelligence}, volume~33, pages 4602--4609, 2019.

\bibitem{pareja2019evolvegcn:graphs}
Aldo Pareja, Giacomo Domeniconi, Jie Chen, Tengfei Ma, Toyotaro Suzumura, Hiroki Kanezashi, Tim Kaler, Tao Schardl, and Charles Leiserson.
\newblock Evolvegcn: Evolving graph convolutional networks for dynamic graphs.
\newblock In {\em Proceedings of the AAAI conference on artificial intelligence}, volume~34, pages 5363--5370, 2020.

\bibitem{poursafaei2022towardsprediction}
Farimah Poursafaei, Shenyang Huang, Kellin Pelrine, and Reihaneh Rabbany.
\newblock Towards better evaluation for dynamic link prediction.
\newblock {\em Advances in Neural Information Processing Systems}, 35:32928--32941, 2022.

\bibitem{radicchi2013abruptnetworks}
Filippo Radicchi and Alex Arenas.
\newblock {Abrupt transition in the structural formation of interconnected networks}.
\newblock {\em Nature Physics}, 9(11):717--720, 7 2013.

\bibitem{rampasek2022recipetransformer}
Ladislav Ramp{\'a}{\v{s}}ek, Michael Galkin, Vijay~Prakash Dwivedi, Anh~Tuan Luu, Guy Wolf, and Dominique Beaini.
\newblock Recipe for a general, powerful, scalable graph transformer.
\newblock {\em Advances in Neural Information Processing Systems}, 35:14501--14515, 2022.

\bibitem{rossi2020tgn:graphs}
Emanuele Rossi, Ben Chamberlain, Fabrizio Frasca, Davide Eynard, Federico Monti, and Michael Bronstein.
\newblock {TGN: Temporal Graph Networks for Deep Learning on Dynamic Graphs}.
\newblock {\em arXiv preprint arXiv:2006.10637}, 6 2020.

\bibitem{sankar2018dysat:networks}
Aravind Sankar, Yanhong Wu, Liang Gou, Wei Zhang, and Hao Yang.
\newblock Dysat: Deep neural representation learning on dynamic graphs via self-attention networks.
\newblock In James Caverlee, Xia~(Ben) Hu, Mounia Lalmas, and Wei Wang, editors, {\em {WSDM} '20: The Thirteenth {ACM} International Conference on Web Search and Data Mining, Houston, TX, USA, February 3-7, 2020}, pages 519--527. {ACM}, 2020.

\bibitem{Seo2016StructuredNetworks}
Youngjoo Seo, Micha{\"e}l Defferrard, Pierre Vandergheynst, and Xavier Bresson.
\newblock Gconv: Structured sequence modeling with graph convolutional recurrent networks.
\newblock In {\em Neural Information Processing: 25th International Conference, ICONIP 2018, Siem Reap, Cambodia, December 13-16, 2018, Proceedings, Part I 25}, pages 362--373. Springer, 2018.

\bibitem{skarding2021foundationssurvey}
Joakim Skarding, Bogdan Gabrys, and Katarzyna Musial.
\newblock {Foundations and Modeling of Dynamic Networks Using Dynamic Graph Neural Networks: A Survey}.
\newblock {\em IEEE Access}, 9:79143--79168, 2021.

\bibitem{topping2021understandingcurvature}
Jake Topping, Francesco~Di Giovanni, Benjamin~Paul Chamberlain, Xiaowen Dong, and Michael~M. Bronstein.
\newblock Understanding over-squashing and bottlenecks on graphs via curvature.
\newblock In {\em The Tenth International Conference on Learning Representations, {ICLR} 2022, Virtual Event, April 25-29, 2022}. OpenReview.net, 2022.

\bibitem{trivedi2018dyrep:graphs}
Rakshit Trivedi, Mehrdad Farajtabar, Prasenjeet Biswal, and Hongyuan Zha.
\newblock {DyRep: Representation Learning over Dynamic Graphs}.
\newblock {\em ICLR 2019}, 3 2018.

\bibitem{valdano2015analyticalnetworks}
Eugenio Valdano, Luca Ferreri, Chiara Poletto, and Vittoria Colizza.
\newblock {Analytical computation of the epidemic threshold on temporal networks}.
\newblock {\em Physical Review X}, 5(2), 2015.

\bibitem{vaswani2017attentionneed}
Ashish Vaswani, Noam Shazeer, Niki Parmar, Jakob Uszkoreit, Llion Jones, Aidan~N. Gomez, Lukasz Kaiser, and Illia Polosukhin.
\newblock {Attention Is All You Need}.
\newblock {\em arXiv preprint arXiv:1706.03762}, 6 2017.

\bibitem{velickovic2017graphnetworks}
Petar Veli{\v{c}}kovi{\'c}, Guillem Cucurull, Arantxa Casanova, Adriana Romero, Pietro Li{\`o}, and Yoshua Bengio.
\newblock Graph attention networks.
\newblock In {\em International Conference on Learning Representations}, 2018.

\bibitem{wang2021tcl:learning}
Lu~Wang, Xiaofu Chang, Shuang Li, Yunfei Chu, Hui Li, Wei Zhang, Xiaofeng He, Le~Song, Jingren Zhou, and Hongxia Yang.
\newblock {TCL: Transformer-based Dynamic Graph Modelling via Contrastive Learning}.
\newblock {\em arXiv preprint arXiv:2105.07944}, 2021.

\bibitem{wang2021inductivewalks}
Yanbang Wang, Yen~Yu Chang, Yunyu Liu, Jure Leskovec, and Pan Li.
\newblock Inductive representation learning in temporal networks via causal anonymous walks.
\newblock In {\em ICLR 2021 - 9th International Conference on Learning Representations}, 1 2021.

\bibitem{wang2022temporalnetwork}
Zehong Wang, Qi~Li, Donghua Yu, and Xiaolong Han.
\newblock Temporal graph transformer for dynamic network.
\newblock In {\em International Conference on Artificial Neural Networks}, pages 694--705. Springer, 2022.

\bibitem{wei2022dgtr:detection}
Siqi Wei, Bin Wu, Aoxue Xiang, Yangfu Zhu, and Chenguang Song.
\newblock {DGTR: Dynamic graph transformer for rumor detection}.
\newblock {\em Frontiers in Research Metrics and Analytics}, 7:1055348, 1 2022.

\bibitem{wu2023simplifyingrepresentations}
Qitian Wu, Wentao Zhao, Chenxiao Yang, Hengrui Zhang, Fan Nie, Haitian Jiang, Yatao Bian, and Junchi Yan.
\newblock Simplifying and empowering transformers for large-graph representations.
\newblock {\em Advances in Neural Information Processing Systems}, 36, 2024.

\bibitem{xu2020tgat:graphs}
Da~Xu, Chuanwei Ruan, Evren Korpeoglu, Sushant Kumar, and Kannan Achan.
\newblock {TGAT: Inductive Representation Learning on Temporal Graphs}.
\newblock {\em ICLR 20}, 2 2020.

\bibitem{xu2018hownetworks}
Keyulu Xu, Weihua Hu, Jure Leskovec, and Stefanie Jegelka.
\newblock How powerful are graph neural networks?
\newblock In {\em 7th International Conference on Learning Representations, {ICLR} 2019, New Orleans, LA, USA, May 6-9, 2019}. OpenReview.net, 2019.

\bibitem{yang2024dynamicsurvey}
Leshanshui Yang, Clement Chatelain, and Sebastien Adam.
\newblock {Dynamic Graph Representation Learning With Neural Networks: A Survey}.
\newblock {\em IEEE Access}, 12:43460--43484, 2024.

\bibitem{yang2021discrete-timespace}
Menglin Yang, Min Zhou, Marcus Kalander, Zengfeng Huang, and Irwin King.
\newblock {Discrete-time Temporal Network Embedding via Implicit Hierarchical Learning in Hyperbolic Space}.
\newblock {\em Proceedings of the ACM SIGKDD International Conference on Knowledge Discovery and Data Mining}, 17:1975--1985, 7 2021.

\bibitem{yang2022time-awareevolution}
Yu~Yang, Hongzhi Yin, Jiannong Cao, Tong Chen, Quoc Viet~Hung Nguyen, Xiaofang Zhou, and Lei Chen.
\newblock {Time-aware Dynamic Graph Embedding for Asynchronous Structural Evolution}.
\newblock {\em IEEE Transactions on Knowledge and Data Engineering}, 35(9):9656--9670, 7 2022.

\bibitem{ying2021graphormer:representation}
Chengxuan Ying, Tianle Cai, Shengjie Luo, Shuxin Zheng, Guolin Ke, Di~He, Yanming Shen, and Tie~Yan Liu.
\newblock {Graphormer: Do Transformers Really Perform Bad for Graph Representation?}
\newblock {\em Advances in Neural Information Processing Systems}, 34:28877--28888, 2021.

\bibitem{ying2018graphpinsage}
Rex Ying, Ruining He, Kaifeng Chen, Pong Eksombatchai, William~L Hamilton, and Jure Leskovec.
\newblock Graph convolutional neural networks for web-scale recommender systems.
\newblock In {\em Proceedings of the 24th ACM SIGKDD international conference on knowledge discovery \& data mining}, pages 974--983, 2018.

\bibitem{you2022roland:graphs}
Jiaxuan You, Tianyu Du, and Jure Leskovec.
\newblock {ROLAND: Graph Learning Framework for Dynamic Graphs}.
\newblock {\em KDD}, 8 2022.

\bibitem{yu2023towardslibrary}
Le~Yu, Leilei Sun, Bowen Du, and Weifeng Lv.
\newblock Towards better dynamic graph learning: New architecture and unified library.
\newblock {\em Advances in Neural Information Processing Systems}, 36:67686--67700, 2023.

\bibitem{zheng2023vdgcnet:model}
Ge~Zheng, Wei~Koong Chai, Jiankang Zhang, and Vasilis Katos.
\newblock {VDGCNeT: A novel network-wide Virtual Dynamic Graph Convolution Neural network and Transformer-based traffic prediction model}.
\newblock {\em Knowledge-Based Systems}, 275:110676, 9 2023.

\end{thebibliography}

\medskip


\appendix
\clearpage
\section{Supra-Laplacian and other positional encoding}

\subsection{Spectral Theory on multi-layer networks}

\label{app:rw_multilayer}

To leverage the benefits of fully-connected spatio-temporal attention across all nodes at multiple timestamps, we encode the spatio-temporal structure by considering a DTDG as a multi-layer graph. For a simple DTDG $\mathcal{G} = \{G_1, G_2, G_3\}$ with a fixed number of nodes, we define the square symmetric supra-adjacency matrix $\bar{A} \in \mathbb{R}^{N \times N}$ as follows:

\begin{equation}
\label{eq:supradj}
    \bar{A} = \begin{pmatrix}
A_1 & I & 0\\
I & A_2 & I \\ 
0 & I & A_3
\end{pmatrix}
\end{equation}

Then, we can utilize the rich spectral properties associated with its supra-Laplacian $\bar{L} = \bar{D} - \bar{A}$. Several studies have analyzed the spectrum of those multi-layer graphs~\cite{cozzomultilayerproperties, dong2014clusteringmanifolds, radicchi2013abruptnetworks}. Especially, \cite{radicchi2013abruptnetworks} demonstrated that $\phi_1$, the Fiedler vector, associated with the second smallest eigenvalue $\lambda_1$, known as the algebraic connectivity or Fiedler value, highlights structural changes between each layer. For a DTDG, this provides valuable information about the graph's dynamics over time. We verified this property experimentally by generating a DTDG containing 3 snapshots of a random Erd\H{o}s-Rényi graph \cite{erddsoni} with 10 nodes each and connecting them temporally according to \cref{eq:supradj} (see illustration on~\cref{fig:supralap}). We then project all nodes of the DTDG onto different vectors associated with eigenvalues $\lambda_i$, with $\lambda_0 \leq \lambda_1 \leq ... \leq \lambda_{\text{max}}$. We observe that projecting onto $\phi_1$ provides dynamic information, while projecting onto $\phi_i$ associated with larger eigenvalues $\lambda_i$ reveals increasingly localized structures. These properties strongly motivate the use of spectral analysis of a multi-layer graph derived from a DTDG to achieve a unified spatio-temporal encoding.

\subsection{Supra-graph construction}

\begin{algorithm}[!ht]
\label{alg:spectrum}
\DontPrintSemicolon
\KwIn{$\mathcal{G}$, $w$,$k$, $t+1$}
\KwOut{$\phi, \Lambda$}
\texttt{adjacencies} $\leftarrow$ [ ] \;
\For{$i$ from $\text{max}(0,t-w)$ to $t$}{
  $A_i \leftarrow \texttt{GetAdjacency}(G_i)$ \;
  $A_i \leftarrow \texttt{RemoveIsolated}(A_i)$ \;
  $A_i \leftarrow \texttt{AddVirtualNode}(A_i)$ \;
  $ \texttt{adjacencies}.\texttt{Append}(A_i)$ \;
}
$\bar{\mathcal{A}} \leftarrow \texttt{BlockDiag}(\texttt{adjacencies})$\tcp*[h]{\cref{eq:supradj}}\;
$\bar{\mathcal{A}} \leftarrow \texttt{AddTempConnection}(\bar{\mathcal{A}})$

\tcp*[h]{Add temporal self-connection only if nodes aren't isolated}\;

$\bar{L} = I - D^{-1/2}\bar{A}D^{-1/2}$\;
$\phi^T\Lambda\phi = \texttt{GetKFirstEigVectors}(\bar{L},k)$ \tcp*[h]{$O(k^2N)$}\;
\caption{Computation of supra-laplacian spectrum}
\end{algorithm}

In practice, when isolated nodes are removed, we obtain a mask of size $N$. This mask helps us identify which nodes are isolated at each time step and determines whether their positional encoding will be $\mathbf{0}^k$ or the projection on the basis of the $k$ eigenvectors. The mask also guides us in adding temporal connections between a node and its past, as isolated nodes do not have temporal connections. In summary, the matrix $\bar{A}$ has a different size from $N \times W$ because we remove isolated nodes and add virtual nodes. The masks help us map the actual indices in $\mathcal{G}$ to the rows in $\bar{A}$.

\begin{figure}
        \centering
        \includegraphics[width=\textwidth]{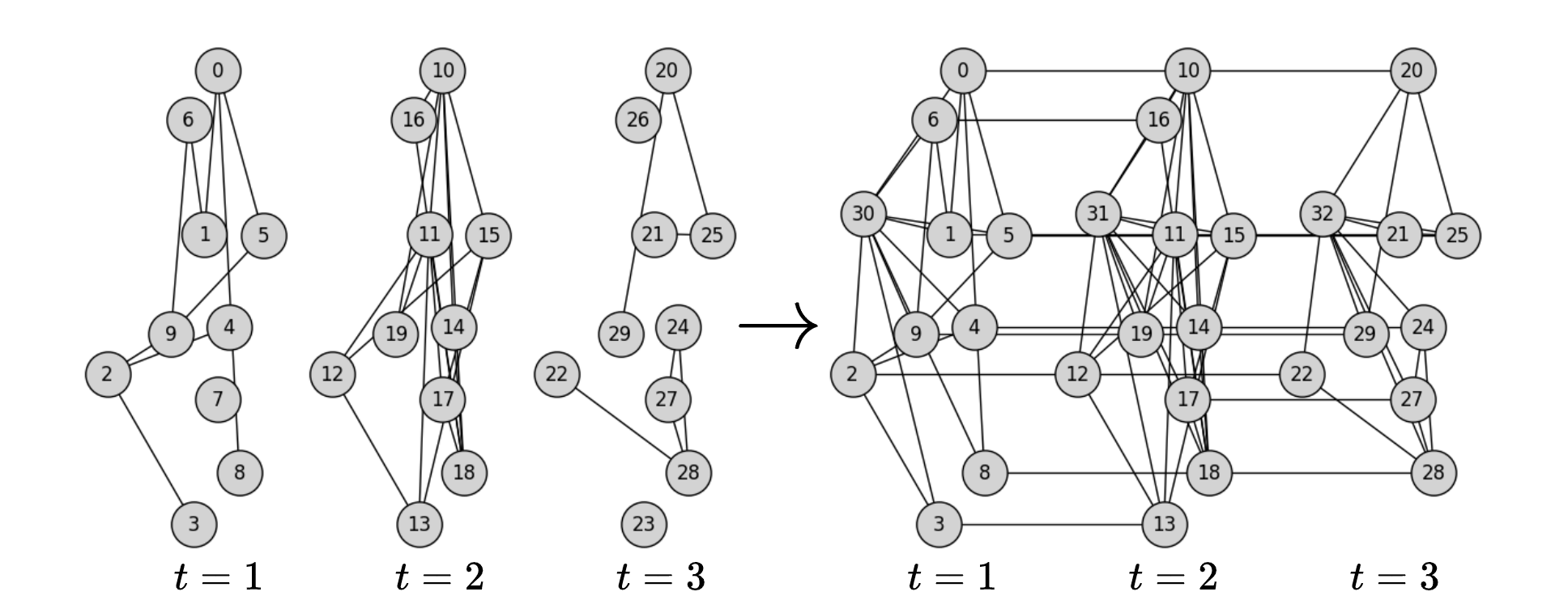}
        \caption{\textcolor{black}{Transformation of a random DTDG into a connected multi-layer network. The \textit{left} side shows independent snapshots with isolated nodes and disconnected clusters. The proposed transformation (\textit{right}) ensures connectivity by introducing temporal edges, removing isolated nodes, and adding a virtual node to connect the clusters within each snapshot.} }
        \label{fig:transformation}
    \end{figure}
\textcolor{black}{In \cref{fig:transformation}, we illustrate the process of transforming a random DTDG into a connected multi-layer network. On the left, we see three independent snapshots, with several isolated nodes (6, 23, and 26) and multiple clusters in the snapshot at $t=3$. The proposed transformation in \ours ensures that the resulting multi-layer graph becomes fully connected by adding temporal connections, removing isolated nodes, and introducing a virtual node to bridge the different clusters within each snapshot. }

\subsection{SupraLaplacian Positional Encoding}
\label{app:supralap}

\textbf{Proof of the positivity of $\lambda_1$ when the graph is connected \cite{Fiedler1973}}\\

\textbf{Theorem 1:} The second smallest eigenvalue, $\lambda_1$ (the Fiedler value), is strictly positive if and only if the graph is connected.

\textbf{Proof 1:} Assume that the graph is not connected. This implies that it can be divided into at least two disjoint connected components without any edges connecting them. For such a graph, it is possible to construct a vector whose entries correspond to these connected components such that the product $\phi_i^T\bar{L}\phi_i=0$, where $\phi_i$ is an eigenvector. This demonstrates that $\lambda_1=0$.
On the contrary, if $\lambda_1 > 0$, the only vector that satisfies $\phi_{i}^T\bar{L}\phi_i=0$ under normal conditions (non-zero $\phi_i$) is the constant vector, indicating that the graph cannot be divided without cutting edges, thus it is connected.

\subsection{Laplacian Positional Encoding}
\label{app:lappe}
\begin{equation}
\label{eq:lappe}
    L_t = I - D_{t}^{-1/2}A_t D_{t}^{-1/2} = \phi_{t}^{T} \Lambda \phi_t
\end{equation}

\begin{equation}
    \label{eq:lpe}
    \text{LapPE}^t_{i} = (\phi^t_{i,1},\phi^t_{i,2},...,\phi^t_{i,d_{\text{pos}}})
\end{equation}

 $L_t$ represents the Laplacian matrix of the graph $G_t$. It is obtained by decomposing the graph as the product of eigenvectors $\phi_t$ and eigenvalues $\Lambda_t$. The Laplacian positional encoding defined in \cref{eq:lpe} provides a unique positional representation of the node $u_{i,t}$ with respect to the $k$ eigenvectors of $G_t$.

\subsection{Unparameterized temporal encoding}
\label{app:timepe}
\begin{equation}
\label{eq:timepe}
    \text{timePE}(t, k) =
    \begin{cases}
        \sin\left(\frac{t}{10000^{(2k/d_{\text{time}})}}\right) & \text{if } k \text{ is even} \\
        \cos\left(\frac{t}{10000^{((2k+1)/d_{\text{time}})}}\right) & \text{if } k \text{ is odd}
    \end{cases}
\end{equation}

In \cref{eq:timepe}, $t$ refers to the $t$-th snapshot of our DTDG $\mathcal{G}$, and $k$ is the dimension in our temporal encoding vector of size $d_{\text{time}}$. This temporal is from~\cite{vaswani2017attentionneed}. To build the ROLAND-GT separate spatio-temporal encoding we concatene the positional encoding LapPE (\cref{eq:lpe}) and the time encoding (\cref{eq:timepe}).

\subsection{GCN Positional Encoding}
\label{app:gcnpe}
We add in our comparison in \cref{fig:supralapvspe}, the GCN positional encoding against our \ours Model. This encoding is derived from a 2-layer GCN as designed by Kipf and Welling \cite{kipf2016semi-supervisednetworks}. This method aggregates the local neighborhood information around a target node with message passing. We use the node embedding as positional encoding to enhance the transformer's awareness of the local structural context. This approach aims to integrate structural insights into the transformer model. It is inspired by the prevalent hybrid architectures combining MP-GNNs and transformers in static Graph Transformers \cite{rampasek2022recipetransformer}. It reflects an evolving trend in graph neural network research, where the strengths of both MP-GNNs in capturing local graph structures and transformers in modeling complex data dependencies are leveraged to enhance model performance on graph-based tasks. However, in our experiments, we found that \ours significantly outperformed the GCN-based positional encoding.

\section{Baselines}
\textbf{Discrete Time Dynamic Graphs Link Prediction models}
We describe the DTDG models from \cite{yang2021discrete-timespace}:
\begin{itemize}
    \item \textbf{GIN and GAT} \cite{kipf2016variationalauto-encoders} : We use static models \cite{xu2018hownetworks,velickovic2017graphnetworks} to showcase the necessity of dynamic model for efficient learning on dynamic graph. GAT is a sparse attention-based model, and GIN is a MP-GNN model design to have a the maximal expressivity of $1-WL$.
    \item \textbf{GRUGCN} \cite{Seo2016StructuredNetworks} : GRUGCN is one of the first discrete dynamic graph GNN models. They introduced the now standard approach which combines a GNN to process the snapshot, and updating embeddings using a temporal model, in their case a GRU.
    
    \item \textbf{EvolveGCN} \cite{pareja2019evolvegcn:graphs}: EvolveGCN is an innovative approach adapting the Graph Convolutional Network (GCN) model for dynamically evolving graphs without relying on node embeddings, effectively capturing the dynamic nature of graph sequences through an RNN to update GCN parameters.
    
    \item \textbf{DySat} \cite{sankar2018dysat:networks} : DySAT use self-attention mechanisms to learn node representations in dynamic graphs. It applies self-attention both structurally and temporally with separate module for time and space, the space module is similar to GAT \cite{velickovic2017graphnetworks}, and the temporal model is a 1-D transformer.
    \item \textbf{VGRNN} \cite{hajiramezanali2019variationalnetworks} : VGRNN  introduce node embedding techniques for dynamic graphs, focusing on variational graph recurrent neural networks to capture temporal dynamics. They employ latent variables for node representation, with SI-VGRNN advancing the model through semi-implicit variational inference for better flexibility. The method is suited for sparse graphs.
    \item \textbf{HTGN} \cite{yang2021discrete-timespace}: They introduce a novel approach for embedding temporal networks through a hyperbolic temporal graph network (HTGN), effectively utilizing hyperbolic space to capture complex, evolving relationships and hierarchical structures in temporal networks. 
    \item \textbf{ROLAND} \cite{you2022roland:graphs}: ROLAND is a generic framework for graph representation learning on DTDG. They allow to efficiently implement any static graph models combine with a RNN-based temporal module.
\end{itemize}
\textbf{Continuous Time Dynamic Graphs Link Prediction models}
We report the description of the CTDG baselines provided in \cite{yu2023towardslibrary}.

\begin{itemize}
    \item \textbf{JODIE} \cite{kumar2019jodie:networks}: Tailored for user-item interaction dynamics within bipartite networks, JODIE utilizes dual recurrent neural networks to refresh user and item states, introducing a projection technique to predict future state trajectories.

    \item \textbf{DyRep} \cite{trivedi2018dyrep:graphs}: Introduces a recurrent mechanism for real-time node state updates, complemented by a temporal attention module to assimilate evolving structural insights of dynamic graphs effectively.

    \item \textbf{TGAT} \cite{xu2020tgat:graphs}: Enhances node representations through the aggregation of temporal-topological neighbor features, leveraging a local self-attention mechanism and time encoding to discern temporal dynamics.

   \item \textbf{TGN} \cite{rossi2020tgn:graphs}: TGN dynamically updates node memories during interactions using a sophisticated mechanism comprising a message function, aggregator, and updater, thereby crafting temporal node representations through an embedding module.

    \item \textbf{CAWN} \cite{wang2021inductivewalks}: Initiates by extracting causal anonymous walks per node to delve into network dynamics and identity correlations, followed by encoding these walks with recurrent neural networks to synthesize comprehensive node representations.

    \item \textbf{EdgeBank} \cite{poursafaei2022towardsprediction}: Adopts a non-parametric, memory-centric strategy for dynamic link prediction, maintaining a repository of interactions for memory updates and utilizing retention-based prediction to distinguish between positive and negative links.

    \item \textbf{TCL} \cite{wang2021tcl:learning} : TCL applies contrastive learning to dynamic graph, using a transformer-based architecture to capture temporal and topological information. It introduces a dual-stream encoder for processing temporal neighborhoods and employs attention mechanisms for semantic inter-dependencies, optimizing through mutual information maximization.
    
    \item \textbf{DyGFormer} \cite{yu2023towardslibrary}: DyGFormer introduces a novel transformer-based architecture for dynamic graph learning, focusing on first-hop interactions to derive node representations. It employs a neighbor co-occurrence encoding scheme to capture node correlations and a patching technique for efficient processing of long temporal sequences. This approach ensures model effectiveness in capturing temporal dependencies and node correlations.
\end{itemize}

\section{Datasets}

\label{app:dts}
\subsection{Datasets description}
\begin{table}[h]
\setlength\tabcolsep{4.5pt}

  \centering
  \footnotesize 
  \captionof{table}{Dataset statistics used in our experiments, with a horizontal bar separating datasets from \cite{yu2023towardslibrary} and datasets from \cite{yang2021discrete-timespace}.}
  \begin{tabular}{l|llll} 
    \toprule
    Datasets & Domains & Nodes & Links & Snapshots \\
    \midrule
    CanParl & Politics & 734 & 74,478 & 14 \\
    USLegis & Politics & 225 & 60,396 & 12  \\
    Flights & Transports & 13,169 & 1,927,145 & 122 \\
    Trade & Economics & 255 & 507,497 & 32  \\
    UNVote & Politics & 201 & 1,035,742 & 72 \\
    Contact & Proximity & 692  & 2,426,279 & 8064 \\
    \midrule
    HepPh & Citations & 15,330 & 976,097 & 36 \\
    AS733 & Router & 6,628  & 13,512 & 30 \\
    Enron & Mail & 184 & 790  & 11 \\
    Colab & Citations & 315 & 943 & 10 \\
    SBM & Synthetic & 1000 & 4,870,863 & 50 \\
    \bottomrule
  \end{tabular}
  \label{tab:datasets}
\end{table}
\begin{itemize}
    \item \textbf{CanParl}: Can. Parl. is a network that tracks how Canadian Members of Parliament (MPs) interacted between 2006 and 2019. Each dot represents an MP, and a line connects them if they both said "yes" to a bill. The line's thickness shows how often one MP supported another with "yes" votes in a year.
    \item \textbf{UsLegis}: USLegis is a Senate co-sponsorship network that records how lawmakers in the US Senate interact socially. The strength of each connection indicates how many times two senators have jointly supported a bill during a specific congressional session
    \item \textbf{Flights}:  Flights is a dynamic flight network that illustrates the changes in air traffic throughout the COVID-19 pandemic. In this network, airports are represented as nodes, and the actual flights are represented as links. The weight of each link signifies the number of daily flights between two airports.
    \item \textbf{Trade}: UNTrade covers the trade in food and agriculture products between 181 nations over a span of more than 30 years. The weight assigned to each link within this dataset reflects the cumulative sum of normalized import or export values for agricultural goods exchanged between two specific countries.
    \item \textbf{UNVote}:  UNVote documents roll-call votes conducted in the United Nations General Assembly. Whenever two nations cast a "yes" vote for an item, the link's weight connecting them is incremented by one.
    \item \textbf{Contact}: Contact dataset provides insights into the evolving physical proximity among approximately 700 university students over the course of a month. Each student is uniquely identified, and links between them indicate their close proximity. The weight assigned to each link reveals the degree of physical proximity between the students
    \item \textbf{Enron}: Enron consists of emails exchanged among 184 Enron employees. Nodes represent employees, and edges indicate email interactions between them. The dataset includes 10 snapshots and does not provide node or edge-specific information
    \item \textbf{Colab}: Colab represents an academic cooperation network, capturing the collaborative efforts of 315 researchers from 2000 to 2009. In this network, each node corresponds to an author, and an edge signifies a co-authorship relationship.
    \item \textbf{HepPH}: HepPh is a citation network focused on high-energy physics phenomenology, sourced from the e-print arXiv website. Within this dataset, each node represents a research paper, while edges symbolize one paper citing another. The dataset encompasses papers published between January 1993 and April 2003, spanning a total of 124 months. 
    \item \textbf{AS733}: AS733 represents an Internet router network dataset, compiled from the University of Oregon Route Views Project. This dataset consists of 733 instances, covering the time period from November 8, 1997, to January 2, 2000, with intervals of 785 days between data points.
    \item \textbf{SBM}: SBM is a synthetic dynamic datasets generated with Stochastic Block Models methods. It contains 1000 nodes and 50 snapshots. We added this datasets, because unlike most of real world datasets, SBM is not a sparse graph. 
\end{itemize}

\subsection{Datasets split}

For the datasets from \cite{yu2023towardslibrary}, we follow the same graph splitting strategy, which means 70\% of the snapshots for training, 15\% for validation, and 15\% for testing. 
We use the same number of snapshots as in HTGN \cite{yang2021discrete-timespace}, the value varies for each dataset (\cref{tab:splitdtdg}).

\begin{table}[!hbt]
    \centering
    \begin{tabular}{c|cccc}
    \toprule
        Datasets & HepPh & AS733  & Enron & Colab \\
        \midrule
        $l$ (number snapshots in \textit{test}) & 6 & 10 & 3 & 3\\
        \bottomrule
    \end{tabular}
    \caption{$l$ represents the number of snapshots in the test dataset. The DTDG is split temporally, following \cite{yang2021discrete-timespace}}
    \label{tab:splitdtdg}
\end{table}

\section{Implementation details and parameters search}
\label{app:param}

For each of our experiments, we used a fixed embedding size of $d=128$, a time window $w=3$, and a single layer of transformer Encoder. Additionally, for the calculation of our positional encoding vectors, we consider that the graph is always undirected. In \cref{tab:searchparam}, we provide the remaining hyperparameters that we adjusted based on the datasets. We selected these datasets by choosing the hyperparameters that yielded the best validation performance in AP.
\texttt{k} is the number of linearly independent eigenvectors we retrieve, it's important to note that $d$ does not increase when \texttt{dim\_pe} grow because $d' = d - \text{k}$. \texttt{nhead\_xa} is the number of head inside the Edge Representation module define in \cref{sec:xa}. \texttt{nhead\_encoder} is the number of head inside \ours \cref{sec:model}, \texttt{dim\_ffn} is the dimension of the feed forward networks in \ours and \texttt{norm\_first} is a condition in \ours to weither or not applying a layer norm before the full attention.

\begin{table}[!ht]
    \centering
    \begin{tabular}{c|c}
    \toprule
        Parameters & Search Range\\
        \midrule
         \texttt{k}& [4,6,10,12,14] \\
         \texttt{nhead\_xa}& [1,2,4,8]\\
         \texttt{nhead\_encoder}& [1,2,4,8]\\
         \texttt{dim\_ffn}& [128,512,1024] \\
         \texttt{norm\_first}& [True,False] \\
         \texttt{learning\_rate}& [0.1,0.01,0.001,0.0001]\\
         \texttt{weight\_decay} & [0,5e-7]\\
         \bottomrule
    \end{tabular}
    \caption{Hyperparameter search range.}
    \label{tab:searchparam}
\end{table}

\section{Experiments: Additionnal results}
\subsection{AP results for DTDG models}
\label{app:dtdg}
We present additional results with Average Precision metrics to evaluate the dynamic link prediction capibility of models. \textbf{\ours} outperforms all other DTDG models across various datasets, achieving the highest average precision (AP) scores. Specifically, \ours surpasses the best-performing model, HTGN, with significant improvements: +1.22 on HepPh, +1.09 on Enron, and +3.33 on SBM. This highlights the effectiveness of our approach in dynamic link prediction tasks.
\begin{table}[!ht]
    \setlength\tabcolsep{0.3pt}
    \caption{Comparison to DTDG models on discrete data. AP}
    \small
    \label{tab:dtdg_full_ap} 
    \centering
    \begin{tabularx}{\textwidth}{  Y | Y | Y | Y | Y | Y | Y }
        \toprule
          Method & HepPh  & AS733 & Enron & Colab & SBM & Avg \\

        \midrule
          GCN  & 73.67 \small{± 1.05}  & 97.11 \small{± 0.01} & 91.00 \small{± 0.73} & 90.17 \small{± 0.25} & 94.57 \small{± 0.30} & 89.30 \small{± 0.47}  \\
          GIN  & 70.55 \small{± 0.84}  & 93.43 \small{± 0.47} & 89.47 \small{± 1.52} & 87.82 \small{± 0.52} & 85.64 \small{± 0.11} & 85.38 \small{± 0.69} \\
          EvolveGCN  & 81.18 \small{± 0.89}  & 95.28 \small{± 0.01} & 92.71 \small{± 0.34} & 87.53 \small{± 0.22} & 92.34 \small{± 0.17} & 89.81 \small{± 0.33} \\
          GRUGCN  & 85.87 \small{± 0.23}  & 96.64 \small{± 0.22} & 93.38 \small{± 0.24} & 87.87 \small{± 0.58} & 91.73 \small{± 0.46} & 91.09 \small{± 0.35} \\
          DySat & 84.47 \small{± 0.23}  & 96.72 \small{± 0.12} & 93.06 \small{± 1.05} & 90.40 \small{± 1.47} & 90.73 \small{± 0.42} & 91.07 \small{± 0.66} \\
          VGRNN & 80.95 \small{± 0.94}  & 96.69 \small{± 0.31} & 93.29 \small{± 0.69} & 87.77 \small{± 0.79} & 90.53 \small{± 0.14} & 89.85 \small{± 0.57 } \\
          HTGN & 89.52 \small{± 0.28}  & \textbf{98.41 }\small{± 0.03} & 94.31 \small{± 0.26} & 91.91 \small{± 0.07} & 94.71 \small{± 0.13} & 93.77 \small{± 0.15} \\
          ROLAND-GT & 82.75 \small{± 0.31}  & 93.66 \small{± 0.14} & 89.86 \small{± 0.29} & 85.03 \small{± 1.96} & 93.62 \small{± 0.28} & 88.98 \small{± 0.59} \\
          \midrule
          \rowcolor{slatecolor}
    \textbf{\ours} & \textbf{90.74} \small{± 0.51}  & {98.16 }\small{± 0.36} & \textbf{95.40 }\small{± 0.29} &\textbf{92.15} \small{± 0.28} & \textbf{98.04 }\small{± 0.29} & \textbf{94.90} \small{± 0.34} \\
          \bottomrule
    \end{tabularx}
\end{table}
\label{app:additionnalexp}
\subsection{Comparison state of the art: Hard Negative Sampling}
\label{app:hardnss}
We present a extensive set of results for our method in comparison to CTDG models in the task of dynamic link prediction on discrete-time dynamic graphs in \cref{tab:ctdg_full_ap} and \cref{tab:ctdg_full_auc}. Here, we emphasize the effectiveness of our model when employing hard historical negative sampling. Historical negative sampling technique (hist) was introduced in \cite{poursafaei2022towardsprediction} to enhance the evaluation of a model's dynamic capability by selecting negatives that occurred in previous time-steps but are not present at the current time for prediction. 
Inductive negative sampling evaluating the capability of models to predict new links that never occured before. 
Our results demonstrate that our model excels at distinguishing hard negative edges compared to other CTDG models, as evidenced by improved performance in both AP and AUC metrics. \ours also consistently outperforms other models using the indudctive (ind) sampling method across multiple datasets, showcasing its superior capability in capturing dynamic graph interactions. Notably, \ours achieves significant improvements on datasets such as USLegis and Trade, demonstrating its robustness and effectiveness in dynamic link prediction tasks.

\subsection{Model Analysis: Additional results}
\begin{figure}[!htb]
\centering 
\begin{minipage}{0.45\textwidth} 
    \centering
    \captionof{table}{Impact of \link module on Dynamic Link Prediction task. ROC-AUC.} 
    \small
    
    \begin{tabular}{c|c|c}
    \toprule
        Datasets &  \ours w/o \link &\cellcolor{slatecolor} \textbf{\ours} \\
        \midrule
        CanParl & 89.45 {± 0.38 } & \cellcolor{slatecolor} \textbf{92.37} {± 0.51 } \\
        USLegis & 93.30 {± 0.29 } & \cellcolor{slatecolor} \textbf{95.80} {± 0.11 }  \\
        Flights & \textbf{99.04} {± 0.61 }& \cellcolor{slatecolor} \textbf{99.07} {± 0.41 } \\
        Trade & 94.01 {± 0.73 } & \cellcolor{slatecolor} \textbf{96.73} {± 0.29 }  \\
        UNVote & 93.56 {± 0.68 } & \cellcolor{slatecolor} \textbf{99.94} {± 0.05 }\\
        Contact & 97.41 {± 0.10} &\cellcolor{slatecolor}  \textbf{98.12} {± 0.37} \\
        \midrule 
        HepPh  & 90.44 {± 1.07 } & \cellcolor{slatecolor} \textbf{93.21} {± 0.37 } \\
        AS733 & 96.84 {± 0.26} & \cellcolor{slatecolor} \textbf{97.46} {± 0.45}\\
        Enron & 90.57 {± 0.27} & \cellcolor{slatecolor} \textbf{96.39} {± 0.18}  \\
        COLAB & 86.34 {± 0.34} & \cellcolor{slatecolor} \textbf{90.84} {± 0.41 } \\
    \bottomrule
    \end{tabular}
    \label{tab:edge_impact}
\end{minipage}%
\hfill
\begin{minipage}{0.5\textwidth} 
    \centering
    \includegraphics[width=\textwidth]{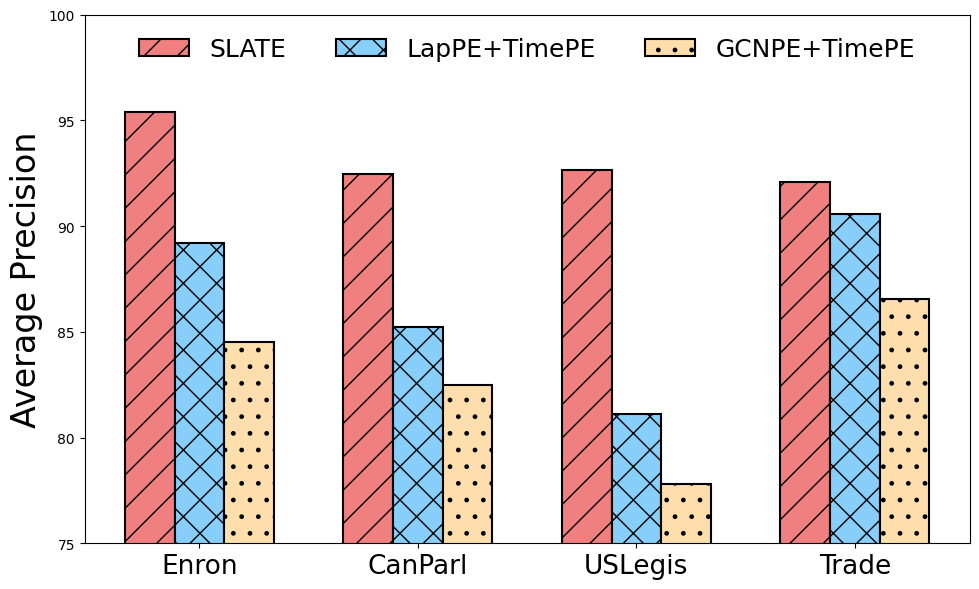}
    \caption{Comparison of \ours encoding against separate structural/positional encoding and time encoding.}
    \label{fig:supralapvspe}
\end{minipage}
\end{figure}

\cref{fig:supralapvspe} provides a comparison between \ours spatio-temporal encoding and separate spatial and temporal encodings, including the Laplacian \cite{dwivedi2021graphrepresentations} (Lap \cref{eq:lappe}) and GCN (\cref{app:gcnpe}) encodings. For calculating the spatial encoding, we selected two common strategies; the first involves using, as we do, the first $k$ eigenvectors of the Laplacian \cite{dwivedi2021graphrepresentations}, but only for the current snapshot. Empirically, we found that the GCN encoding did not yield satisfying results, in contrast to the hybrid architecture strategies widely used for static Graph Transformers \cite{rampasek2022recipetransformer}.

 We show in \cref{tab:edge_impact} \ours, with its cross-attention mechanism for edge representation, significantly enhances the predictive accuracy of the \ours w/o \link model. We show improvement across various datasets, further emphasizing the importance of modeling temporal interactions explicitly, we gain for example +6.4 points on UNVote, +2.6 points on USLegis and +5.8 points on Enron. \ours's ability to capture intricate dynamics between two nodes across time dimensions results in substantial performance gains, making it a valuable addition to our model architecture.

\textbf{Impact of the time-pooling function.} In \cref{tab:timepool}, we present the performance of the time-pooling function  used in \cref{sec:xa}, across the US Legis, UN Vote, and Trade datasets, with the time window set to $w=3$. Using $k=3$ corresponds to averaging over all snapshots within the window, whereas $k=1$ focuses exclusively on the last element of $E_{uv}$. The results indicate that averaging (mean pooling) consistently outperforms max pooling, irrespective of the $k$ value. For our primary analysis, we therefore adopt $k=3$.
\begin{table}[!ht]
    \small
    \caption{Comparison of \ours for several time pooling methods  (random sampling), on USLegis, UNVote and Trade. }
    
    \label{tab:timepool} 
    \centering

    \begin{tabular}{l | c c | c c | c c  }
        \toprule
         \multirow{2}{*}{Pool} & \multicolumn{2}{c}{USLegis} & \multicolumn{2}{c}{UNvote} & \multicolumn{2}{c}{Trade}  \\  
         &  AUC & AP & AUC & AP & AUC & AP\\
         \midrule
         Max &  93.03 & 88.68 &87.92 & 87.72 & 93.37 & 93.32\\
         Avg. $k=3$ &  94.50 & 89.67 &\textbf{99.72} & \textbf{99.75} & 96.71 & 96.88 \\
         Avg. $k=2$ & \textbf{95.35} & \textbf{92.17} & 99.69 & 99.67 & 96.76 & 96.93 \\
         Avg. $k=1$ & \text{94.67} & \text{91.28} & 99.59 & 99.44 & \textbf{96.78} & \textbf{96.97}\\
         \bottomrule
    \end{tabular}
\end{table}

\textcolor{black}{\textbf{Detailed analysis of the DTDG-to-multi-layer transformation in SLATE}}
\textcolor{black}{We provide a closer examination of the performance of SLATE under various transformations applied to the DTDG during its conversion into a multi-layer graph. The \cref{tab:isolated_vs_slate} demonstrates the negative effect of retaining isolated nodes, which leads to a significant drop in performance on both the Colab and USLegis datasets. By removing these nodes and focusing on the spectrum associated with the first non-zero eigenvalue, SLATE achieves a substantial performance improvement.}

\textcolor{black}{The \cref{tab:with_wo_vn} highlights the importance of introducing a virtual node (VN) that connects clusters within each snapshot. Without the VN, the model underperforms, as shown in the results for the Enron dataset. This confirms that each transformation step, from removing isolated nodes to adding temporal connections and VNs, plays a critical role in enhancing the quality of the spatio-temporal encoding.}

\begin{table}[!ht]
    \centering
    \begin{minipage}{0.45\textwidth}
        \centering
        \begin{tabular}{ccc}
        \toprule
             Models & Colab & USLegis \\
             \midrule
             \ours with isolated nodes & 86.73 & 66.57\\
             \rowcolor{slatecolor} \ours & \textbf{90.84} & \textbf{95.80}\\
             \bottomrule
        \end{tabular}
        \caption{Performance impact in AUC of keeping isolated nodes on Colab and USLegis datasets. Removing isolated nodes and focusing on the first non-zero eigenvalue spectrum leads to a significant performance boost.}
        \label{tab:isolated_vs_slate}
    \end{minipage}%
    \hspace{0.05\textwidth} 
    \begin{minipage}{0.45\textwidth}
        \centering
        \begin{tabular}{ccc}
        \toprule
             Models & AP & AUC \\
             \midrule
             \ours w/o VN & 93.74 & 95.18\\
             \rowcolor{slatecolor} \ours & \textbf{95.40} & \textbf{96.39}\\
             \bottomrule
        \end{tabular}
        \caption{Effect of introducing a virtual node (VN) on the Enron dataset. The addition of the VN improves SLATE's performance in terms of both AP and AUC.}
        \label{tab:with_wo_vn}
    \end{minipage}
\end{table}

\textcolor{black}{\textbf{AUC : Impact of time window on multiple models}}
\textcolor{black}{The analysis in \cref{tab:tw_models} demonstrates that the impact of the time window on model performance is consistent across different types of models, including our transformer-based approach and two MP-GNNs (EGCN and DySAT). Interestingly, we observe that a relatively short time window produces optimal results for all models on the UNVote dataset, which spans 72 snapshots. Specifically, both EGCN and DySAT achieve their highest AUC with $W=4$, while \ours achieves peak performance at $W=2$. This indicates that capturing spatio-temporal dynamics does not necessarily require long temporal windows, and in fact, shorter windows can often lead to better performance by focusing on more immediate temporal interactions.}
\begin{table}[!ht]
    \centering
    \begin{tabular}{l|c|cccccc}
         \toprule
         Model & Nb param. & $W=1$ & $W=2$ & $W=3$& $W=4$ & $W=5$ & $W=\infty$ \\
         \midrule
         
         EGCN & 1.8M &86.96 & 86.48 & 86.74 & \textbf{87.66} & 85.26 & 86.74\\
         DySAT & 1.8M & 83.93 & 81.90 & 86.15 & \textbf{88.71} & 80.08 & 77.04\\
          \rowcolor{slatecolor} \textbf{\ours} & 2.1M & 96.68 & \textbf{99.73} & 98.74 & 95.90 & 95.79 & 92.24\\
         \bottomrule
    \end{tabular}
    \caption{Effect of time window size on AUC for different models. Shorter windows provide optimal results across all models.}
    \label{tab:tw_models}
\end{table}

\subsection{More qualitative analysis}
\begin{figure}[!ht]
        \centering
        \includegraphics[width=0.85\textwidth]{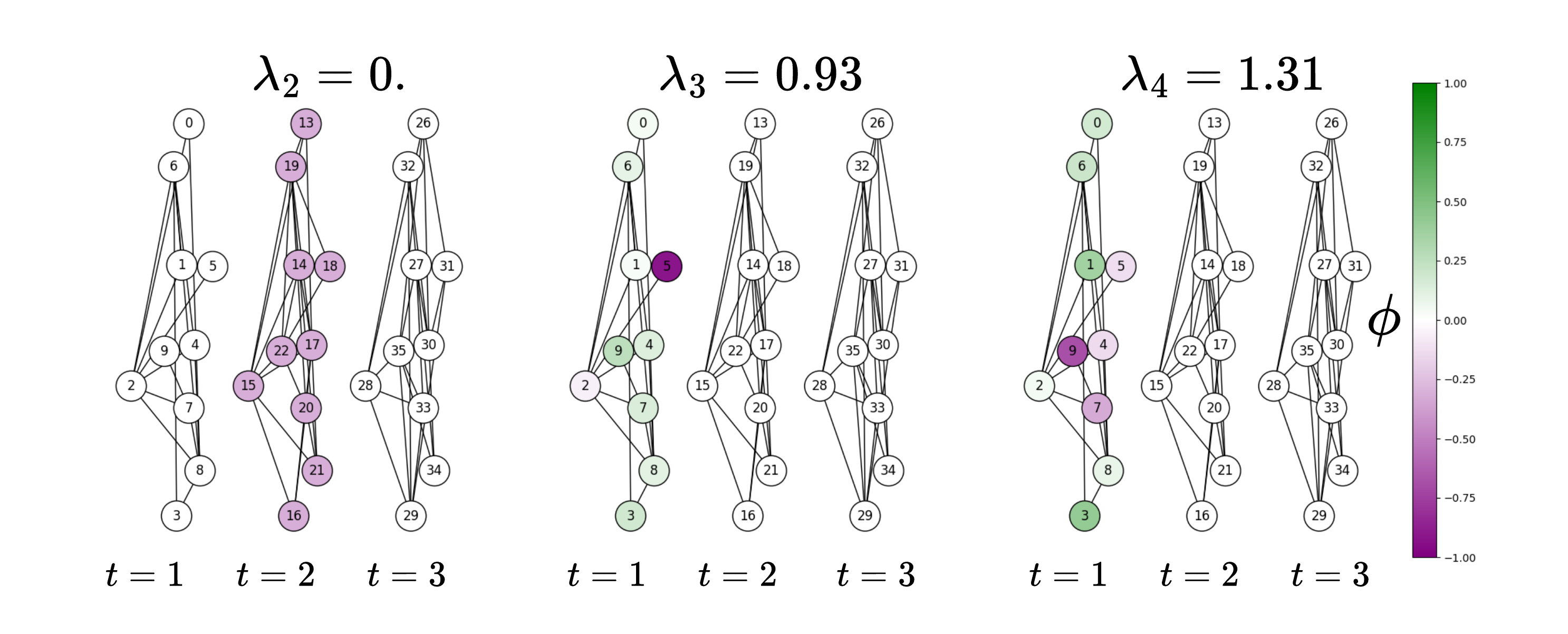}
        \caption{\textcolor{black}{Effect of missing temporal connections in a DTDG. Without temporal edges, the figure illustrates that the projections are purely spatial, and the three snapshots remain independent, with no spatio-temporal interaction captured.}}
        \label{fig:quali_no_temp}
    \end{figure}

\begin{figure}[!ht]
        \centering
        \includegraphics[width=0.85\textwidth]{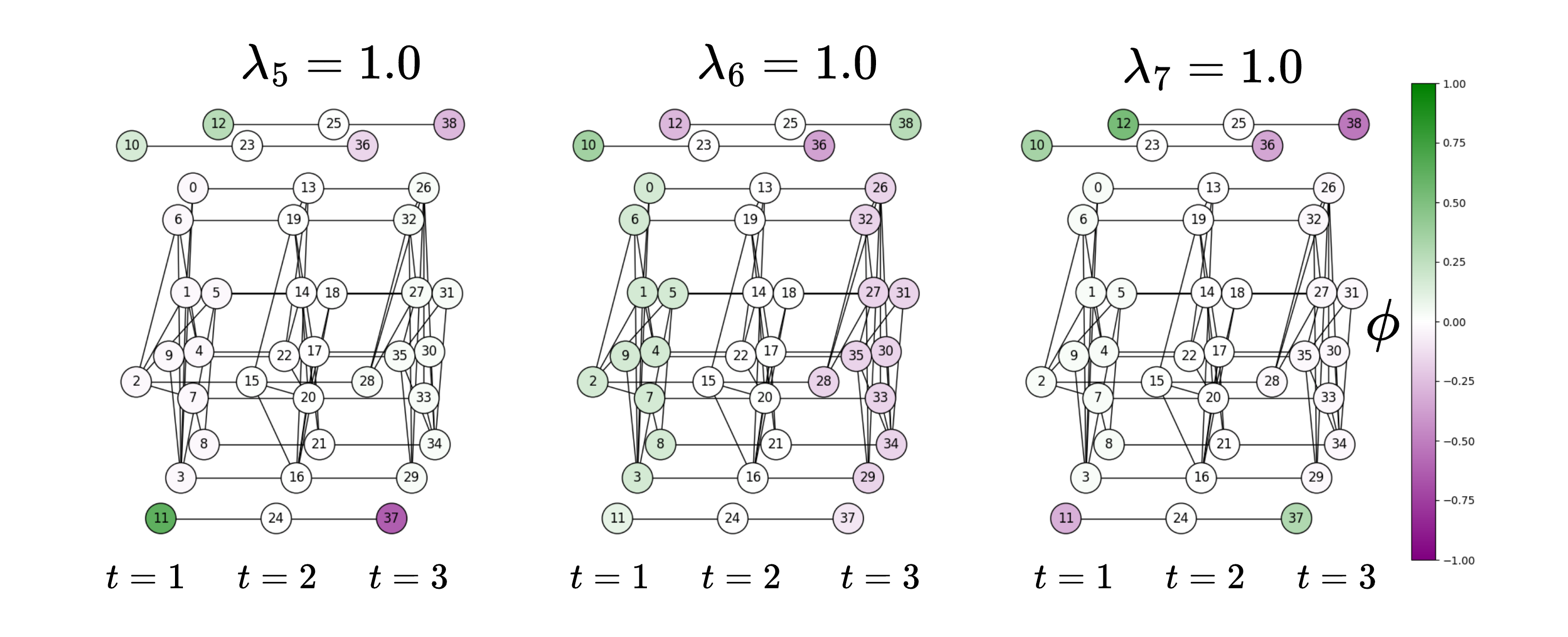}
        \caption{\textcolor{black}{Effect of retaining isolated nodes in a DTDG with added temporal connections. The figure shows that keeping isolated nodes results in multiple disconnected components, where many projections focus on these nodes, obscuring the overall spatio-temporal structure of the graph.}}
        \label{fig:quali_isolated}
    \end{figure}

\textcolor{black}{We conduct a fine-grained analysis of the impact of not processing the DTDG correctly. \cref{fig:quali_no_temp} demonstrates that without temporal connections, the result is purely spatial projections with no spatio-temporal information, as the three snapshots remain independent. \cref{fig:quali_isolated} illustrates the effect of retaining isolated nodes while adding temporal connections. Keeping these nodes leads to multiple disconnected components in the graph, where many projections focus solely on the isolated nodes, neglecting the core structure of the DTDG. This issue is further intensified by the fact that we consider only the $k$ eigenvectors associated with the first non-zero eigenvalue, limiting the ability to capture the full spatio-temporal dynamics.}

\subsection{Scalability}
\textcolor{black}{In \cref{tab:efficiency}, we demonstrate that Performer \cite{choromanski2022rethinkingattentionperformers} significantly reduces memory consumption and speeds up training time per epoch. Moreover, as shown in \cref{tab:performer_vs_transformer}, using Performer an efficient approximation of the attention matrix with linear complexity—does not significantly degrade the results compared to the standard Transformer encoder. Performer is a highly advantageous solution for scaling to larger graphs while maintaining the benefits of dynamic graph transformers. Its linear complexity allows it to handle larger datasets efficiently, without sacrificing performance.}

\begin{table}[!ht]
    \centering
    \begin{tabular}{l|ccc}
    \toprule
         Models & AS733 & USLegis & UNtrade\\
    \midrule
         SLATE-Transformer & 97.46 ± 0.45 & 95.80 ± 0.11 & 96.73 ± 0.29\\
         SLATE-Performer & 95.39 ± 0.61 & 95.14 ± 0.84 & 96.21 ± 0.77\\
    \bottomrule
    \end{tabular}
    \caption{AUC performance comparison between \ours using a standard Transformer encoder (\cite{  vaswani2017attentionneed}) and a Performer encoder (\cite{choromanski2022rethinkingattentionperformers})}
    \label{tab:performer_vs_transformer}
\end{table}

\clearpage

\begin{table*}[!ht]
    \setlength\tabcolsep{0.3pt}
    \caption{Comparison to CTDG models on discrete data. ROC-AUC}
    
    \label{tab:ctdg_full_auc} 
    \centering
    \begin{tabularx}{\textwidth}{ c| Y | Y | Y | Y | Y | Y | Y }
        \toprule
         NSS & Method & CanParl  & USLegis & Flights & Trade & UNVote & Contact\\

        \midrule
          \multirow{11}{*}{rns} & JODIE &  78.21 \small{± 0.23}  & 82.85 \small{± 1.07} &96.21 \small{± 1.42} &69.62 \small{± 0.44} &68.53 \small{± 0.95} &96.66 \small{± 0.89}\\
          &DyREP  & 73.35 \small{± 3.67}  & 82.28\small{ ± 0.32} & 95.95 \small{± 0.62} & 67.44 \small{± 0.83} & 67.18 \small{± 1.04} & 96.48 \small{± 0.14}\\
          &TGAT  & 75.69 \small{± 0.78} &  75.84 \small{± 1.99} & 94.13 \small{± 0.17} &64.01 \small{± 0.12} & 52.83 \small{± 1.12}& 96.95 \small{± 0.08}\\
          &TGN  & 76.99 \small{± 1.80} & \underline{83.34 \small{± 0.43}} &98.22 \small{± 0.13} &69.10 \small{± 1.67} & \underline{69.71 \small{± 2.65}}& 97.54 \small{± 0.35}\\
          &CAWN  & 75.70\small{ ± 3.27} & 77.16 \small{± 0.39} &98.45 \small{± 0.01} & 68.54 \small{± 0.18}&53.09 \small{± 0.22} & 89.99 \small{± 0.34}\\
          &EdgeBank  & 64.14 \small{± 0.00}  & 62.57 \small{± 0.00} &90.23\small{ ± 0.00} & 66.75 \small{± 0.00} & 62.97 \small{± 0.00} & 94.34 \small{± 0.00}\\
          &TCL  & 72.46 \small{± 3.23} & 76.27 \small{± 0.63} &91.21 \small{± 0.02} &64.72 \small{± 0.05} &51.88 \small{± 0.36} & 94.15 \small{± 0.09}\\
          &GraphMixer  & 83.17 \small{± 0.53} & 76.96 \small{± 0.79}  & 91.13 \small{± 0.01}& 65.52 \small{± 0.51}&52.46 \small{± 0.27} & 93.94 \small{± 0.02}\\
          &DyGformer  & \textbf{97.76 \small{± 0.41}} &  77.90 \small{± 0.58} & \underline{98.93 \small{± 0.01}}&\underline{70.20 \small{± 1.44}} & 57.12 \small{± 0.62} & \textbf{98.53 \small{± 0.01}}\\
          \cmidrule{2-8}
          & \cellcolor{slatecolor} \textbf{\ours}   & \cellcolor{slatecolor} \underline{92.37 \small{± 0.51}}& \cellcolor{slatecolor} \textbf{95.80 \small{± 0.11}} & \cellcolor{slatecolor} \textbf{99.07 ± 0.41 }&\cellcolor{slatecolor} \textbf{96.73 ± 0.29} & \cellcolor{slatecolor}\textbf{99.94 ± 0.05} & \cellcolor{slatecolor} \underline{98.12 ± 0.37}\\
          
         \bottomrule
         \multirow{11}{*}{hist} & JODIE & 62.44 \small{± 1.11} & 67.47 \small{± 6.40} & 68.97 \small{± 1.87}&68.92 \small{± 1.40} &76.84\small{ ± 1.01} & \underline{96.35 \small{± 0.92}}\\
          &DyREP  &70.16 \small{± 1.70} &\small{91.44 ± 1.18} & 69.43 \small{± 0.90} &64.36 \small{± 1.40} & 74.72 \small{± 1.43}& 96.00 \small{± 0.23}\\
          &TGAT  & 70.86 \small{± 0.94}&73.47 \small{± 5.25} &72.20 \small{± 0.16} &60.37 \small{± 0.68} &53.95 \small{± 3.15} & 95.39 \small{± 0.43}\\
          &TGN  & 73.23 \small{± 3.08}& 83.53 \small{± 4.53} &68.39 \small{± 0.95} & 63.93 \small{± 5.41}&73.40 \small{± 5.20} & 93.76 \small{± 1.29}\\
          &CAWN  &72.06 \small{± 3.94} & 78.62 \small{± 7.46} & 66.11 \small{± 0.71} & 63.09 \small{± 0.74}&51.27 \small{± 0.33} & 83.06 \small{± 0.32}\\
          &EdgeBank  &63.04 \small{± 0.00} & 67.41 \small{± 0.00} & \underline{74.64 \small{± 0.00}}&\underline{86.61 ± 0.00} &\underline{89.62 \small{± 0.00}} & 92.17 \small{± 0.00}\\
          &TCL  & 69.95 \small{± 3.70}& 83.97 \small{± 3.71} & 70.57 \small{± 0.18} & 61.43 \small{± 1.04}& 52.29 \small{± 2.39}& 93.34 \small{ ± 0.19}\\
          &GraphMixer  & 79.03 \small{± 1.01}&85.17 \small{± 0.70} & 70.37 \small{± 0.23}& 63.20 \small{± 1.54}&52.61 \small{± 1.44} &93.14 \small{± 0.34}\\
          &DyGformer  & \textbf{97.61 ± 0.40}& \underline{90.77 ± 1.96}& 68.09 \small{± 0.43}& 73.86 \small{± 1.13}&64.27 \small{± 1.78} &\textbf{97.17 ± 0.05}\\
          \cmidrule{2-8}
          &\cellcolor{slatecolor} \textbf{\ours}   & \cellcolor{slatecolor} \underline{88.71 \small{± 0.43}} & \cellcolor{slatecolor} \textbf{\underline{90.69 \small{± 0.50}}} & \cellcolor{slatecolor} \textbf{76.83 ± 0.69} & \cellcolor{slatecolor} \textbf{92.14 ± 0.38} & \cellcolor{slatecolor} \textbf{98.62 ± 0.49}  & \cellcolor{slatecolor} 94.29 \small{± 0.09}\\
          \bottomrule
          \multirow{11}{*}{ind} &JODIE & 52.88 ± 0.80 & 59.05 ± 5.52  & 69.99 ± 3.10 & 66.82 ± 1.27 &  \underline{73.73} ± 1.61 & \underline{94.47} ± 1.08 \\
          &DyREP  & 63.53 ± 0.65 & \underline{89.44} ± 0.71 &71.13 ± 1.55& 65.60 ± 1.28 & 72.80 ± 2.16 & 94.23 ± 0.18  \\
          &TGAT  & 72.47 ± 1.18 & 71.62 ± 5.42 & 73.47 ± 0.18 & 66.13 ± 0.78 & 53.04 ± 2.58 & 94.10 ± 0.41 \\
          &TGN  & 69.57 ± 2.81 & 78.12 ± 4.46  & 71.63 ± 1.72 & 66.37 ± 5.39 & 72.69 ± 3.72 & 91.64 ± 1.72  \\
          &CAWN  & 72.93 ± 1.78 & 76.45 ± 7.02  & 69.70 ± 0.75 & 71.73 ± 0.74  &  52.75 ± 0.90 & 87.68 ± 0.24 \\
          &EdgeBank  & 61.41 ± 0.00 &  68.66 ± 0.00 & \textbf{81.10} ± 0.00  & \underline{74.20} ± 0.00  & 72.85 ± 0.00 & 85.87 ± 0.00\\
          &TCL  & 69.47 ± 2.12 & 82.54 ± 3.91 & 72.54 ± 0.19 & 67.80 ± 1.21 & 52.02 ± 1.64   & 91.23 ± 0.19  \\
          &GraphMixer  & 70.52 ± 0.94 & 84.22 ± 0.91 & 72.21 ± 0.21 & 66.53 ± 1.22  & 51.89 ± 0.74 & 90.96 ± 0.27  \\
          &DyGformer  & \textbf{96.70} ± 0.59 & 87.96 ± 1.80 & 69.53 ± 1.17 & 62.56 ± 1.51 & 53.37 ± 1.26 & \textbf{95.01} ± 0.15 \\
          \cmidrule{2-8}
          &\cellcolor{slatecolor} \textbf{\ours} & \cellcolor{slatecolor} \underline{93.74} ± 0.08 &\cellcolor{slatecolor} \textbf{90.23} ± 0.29 & \cellcolor{slatecolor} \underline{76.98} ± 1.64 &\cellcolor{slatecolor} \textbf{91.45} ± 0.39 & \cellcolor{slatecolor}  \textbf{92.78} ± 0.06  &  \cellcolor{slatecolor} {94.03} ± 0.43\\
          \bottomrule
    \end{tabularx}
\end{table*}
\begin{table*}[!ht]
    \setlength\tabcolsep{0.3pt}
    \caption{Comparison to CTDG models on discrete data. Average Precision .}
    
    \label{tab:ctdg_full_ap}  
    \centering
    \begin{tabularx}{\textwidth}{c |Y | Y | Y | Y | Y | Y | Y }
        \toprule
         NSS & Method & CanParl  & USLegis & Flights & Trade & UNVote & Contact\\
        \midrule
          \multirow{12}{*}{rns} &JODIE & 69.26 ± 0.31  & 75.05 ± 1.52 & 95.60 ± 1.73 & 64.94 ± 0.31 & 63.91 ± 0.81 & 95.31 ± 1.33\\
          &DyREP  &66.54 \small{± 2.76}  &75.34 \small{± 0.39}& 95.29 \small{± 0.72} & 63.21 \small{± 0.93} &62.81 \small{± 0.80} & 95.98 \small{± 0.15}\\
          &TGAT  & 70.73 ± 0.72 &68.52 ± 3.16 &94.03 ± 0.18 & 61.47 ± 0.18& 52.21 ± 0.98 & 96.28 ± 0.09\\
          &TGN  &70.88 ± 2.34 & \underline{75.99 ± 0.58} &97.95 ± 0.14 & 65.03 ± 1.37 & \underline{65.72 ± 2.17} & 96.89 ± 0.56\\
          &CAWN  &69.82 ± 2.34 & 70.58 ± 0.48 &98.51 ± 0.01 &65.39 ± 0.12 &52.84 ± 0.10 & 90.26 ± 0.28\\
          &EdgeBank  & 64.55 ± 0.00  &58.39 ± 0.00 & 89.35 ± 0.00 &60.41 ± 0.00 & 58.49 ± 0.00 & 92.58 ± 0.00\\
          &TCL  & 68.67 ± 2.67  & 69.59 ± 0.48 & 91.23 ± 0.02 & 62.21 ± 0.03 & 51.90 ± 0.30& 92.44 ± 0.12\\
          &GraphMixer  & 77.04 ± 0.46 & 70.74 ± 1.02 & 90.99 ± 0.05 & 62.61 ± 0.27 &52.11 ± 0.16 & 91.92 ± 0.03\\
          &DyGformer  & \textbf{97.36 ± 0.45} & 71.11 ± 0.59 & \textbf{98.91 ± 0.01} &\underline{66.46 ± 1.29} & 55.55 ± 0.42& \textbf{98.29 ± 0.01}\\
          \cmidrule{2-8}
          &\cellcolor{slatecolor} \textbf{\ours}  & \cellcolor{slatecolor} \underline{92.44 \small{± 0.25}} & \cellcolor{slatecolor} \textbf{92.66 \small{± 0.41}} & \cellcolor{slatecolor} \underline{98.61 \small{± 0.44}} & \cellcolor{slatecolor} \textbf{96.91 \small{± 0.23}} &\cellcolor{slatecolor} \textbf{99.91 \small{± 0.09}} & \cellcolor{slatecolor} \underline{97.68 \small{± 0.13}}\\
          
         \bottomrule
         \multirow{11}{*}{hist} &JODIE & 51.79 ± 0.63 &51.71 ± 5.76 & 66.48 ± 2.59 &61.39 ± 1.83 &70.02 ± 0.81 & 95.31 ± 2.13\\
          &DyREP  &63.31 ± 1.23 & \textbf{86.88 ± 2.25} & 67.61 ± 0.99& 59.19 ± 1.07&69.30 ± 1.12 & \underline{96.39 ± 0.20}\\
          &TGAT  & 67.13 ± 0.84 &62.14 ± 6.60 &\underline{72.38 ± 0.18} &55.74 ± 0.91 &52.96 ± 2.14 & 96.05 ± 0.52\\
          &TGN  & 68.42 ± 3.07& 74.00 ± 7.57&66.70 ± 1.64 &58.44 ± 5.51 & 69.37 ± 3.93& 93.05 ± 2.35\\
          &CAWN  & 66.53 ± 2.77&68.82 ± 8.23 & 64.72 ± 0.97&55.71 ± 0.38 &51.26 ± 0.04 & 84.16 ± 0.49\\
          &EdgeBank  &63.84 ± 0.00 & 63.22 ± 0.00&70.53 ± 0.00 & \underline{81.32 ± 0.00} & \underline{84.89 ± 0.00}& 88.81 ± 0.00\\
          &TCL  & 65.93 ± 3.00&80.53 ± 3.95 &70.68 ± 0.24 & 55.90 ± 1.17&52.30 ± 2.35 & 93.86 ± 0.21\\
          &GraphMixer  & 74.34 ± 0.87 &81.65 ± 1.02 &71.47 ± 0.26 & 57.05 ± 1.22& 51.20 ± 1.60& 93.36 ± 0.41\\
          &DyGformer  &\textbf{97.00 \small{± 0.31}} &\underline{85.30 \small{± 3.88}} &66.59 \small{± 0.49} & 64.41 \small{± 1.40}&60.84 \small{± 1.58} & \textbf{97.57 \small{± 0.06}}\\
          \cmidrule{2-8}
          &\cellcolor{slatecolor} \textbf{\ours}  &\cellcolor{slatecolor} \underline{84.38 \small{± 0.81}} & \cellcolor{slatecolor} 83.53 \small{± 1.64}& \cellcolor{slatecolor} \textbf{75.09 \small{± 1.17}}& \cellcolor{slatecolor} \textbf{84.05 \small{± 0.98}} & \cellcolor{slatecolor}\textbf{96.85 \small{± 0.27}} & \cellcolor{slatecolor} 93.58  \small{± 0.16} \\
          
         \bottomrule
         \multirow{11}{*}{ind} &JODIE & 48.42 ± 0.66 & 50.27 ± 5.13 & 69.07 ± 4.02 & 60.42 ± 1.48 &\underline{67.79} ± 1.46   & 93.43 ± 1.78 \\
          &DyREP  & 58.61 ± 0.86 &\underline{83.44} ± 1.16 & 70.57 ± 1.82 &60.19 ± 1.24 &67.53 ± 1.98 &94.18 ± 0.10\\
          &TGAT  & 68.82 ± 1.21 & 61.91 ± 5.82 & 75.48 ± 0.26 &60.61 ± 1.24&52.89 ± 1.61  & 94.35 ± 0.48 \\
          &TGN  & 65.34 ± 2.87 & 67.57 ± 6.47 & 71.09 ± 2.72 & 61.04 ± 6.01 &67.63 ± 2.67  & 90.18 ± 3.28 \\
          &CAWN  & 67.75 ± 1.00 & 65.81 ± 8.52 & 69.18 ± 1.52 &62.54 ± 0.67 &52.19 ± 0.34  & 89.31 ± 0.27 \\
          &EdgeBank  & 62.16 ± 0.00 & 64.74 ± 0.00 & \textbf{81.08} ± 0.00  & \underline{72.97} ± 0.00 & 66.30 ± 0.00 & 85.20 ± 0.00 \\
          &TCL  & 65.85 ± 1.75 & 78.15 ± 3.34 & 74.62 ± 0.18 & 61.06 ± 1.74 & 50.62 ± 0.82 & 91.35 ± 0.21 \\
          &GraphMixer  & 69.48 ± 0.63   & 79.63 ± 0.84 & 74.87 ± 0.21 & 60.15 ± 1.29 & 51.60 ± 0.73  & 90.87 ± 0.35  \\
          &DyGformer  & \textbf{95.44 ± 0.57 } & 81.25 ± 3.62 &  70.92 ± 1.78  & 55.79 ± 1.02 & 51.91 ± 0.84 &  \textbf{94.75} ± 0.28\\
          \cmidrule{2-8}
          &\cellcolor{slatecolor} \textbf{\ours}  & \cellcolor{slatecolor}\underline{ 93.42} ± 0.75 & \cellcolor{slatecolor} \textbf{95.21} ± 0.51 & \cellcolor{slatecolor} \underline{79.03} ± 0.95 & \cellcolor{slatecolor} \textbf{92.87} ± 0.62 & \cellcolor{slatecolor} \textbf{93.74} ± 0.29 & \cellcolor{slatecolor} \underline{94.52} ± 0.86 \\
          \bottomrule

    \end{tabularx}
\end{table*}
\clearpage
\section*{NeurIPS Paper Checklist}

\begin{enumerate}

\item {\bf Claims}
    \item[] Question: Do the main claims made in the abstract and introduction accurately reflect the paper's contributions and scope?
    \item[] Answer: \answerYes{} 
    \item[] Justification: Our paper focuses on a unified spatio-temporal
      encoding based on the spectrum of the supra-Laplacian, as developed in
      \cref{sec:supralaplacian}. We also introduce a fully-connected architecture utilizing this spatio-temporal encoding \cref{sec:full_attention_transformer} for the task of link prediction \cref{sec:xa}. Each of these claims is validated in \cref{tab:impact_components}, as well as the claim of better \ours performance against state-of-the-art methods in \cref{tab:ctdg_main_auc,tab:dtdg_main_auc}.
    \item[] Guidelines:
    \begin{itemize}
        \item The answer NA means that the abstract and introduction do not include the claims made in the paper.
        \item The abstract and/or introduction should clearly state the claims made, including the contributions made in the paper and important assumptions and limitations. A No or NA answer to this question will not be perceived well by the reviewers. 
        \item The claims made should match theoretical and experimental results, and reflect how much the results can be expected to generalize to other settings. 
        \item It is fine to include aspirational goals as motivation as long as it is clear that these goals are not attained by the paper. 
    \end{itemize}

\item {\bf Limitations}
    \item[] Question: Does the paper discuss the limitations of the work performed by the authors?
    \item[] Answer: \answerYes{}  
    \item[] Justification: We discuss the limitations of \ours in the conclusion \cref{sec:conclu}, where we list multiple negative points of our work and suggest possible improvements, particularly in terms of better scalability and evaluating \ours on other graph or node-based tasks.
    \item[] Guidelines:
    \begin{itemize}
        \item The answer NA means that the paper has no limitation while the answer No means that the paper has limitations, but those are not discussed in the paper. 
        \item The authors are encouraged to create a separate "Limitations" section in their paper.
        \item The paper should point out any strong assumptions and how robust the results are to violations of these assumptions (e.g., independence assumptions, noiseless settings, model well-specification, asymptotic approximations only holding locally). The authors should reflect on how these assumptions might be violated in practice and what the implications would be.
        \item The authors should reflect on the scope of the claims made, e.g., if the approach was only tested on a few datasets or with a few runs. In general, empirical results often depend on implicit assumptions, which should be articulated.
        \item The authors should reflect on the factors that influence the performance of the approach. For example, a facial recognition algorithm may perform poorly when image resolution is low or images are taken in low lighting. Or a speech-to-text system might not be used reliably to provide closed captions for online lectures because it fails to handle technical jargon.
        \item The authors should discuss the computational efficiency of the proposed algorithms and how they scale with dataset size.
        \item If applicable, the authors should discuss possible limitations of their approach to address problems of privacy and fairness.
        \item While the authors might fear that complete honesty about limitations might be used by reviewers as grounds for rejection, a worse outcome might be that reviewers discover limitations that aren't acknowledged in the paper. The authors should use their best judgment and recognize that individual actions in favor of transparency play an important role in developing norms that preserve the integrity of the community. Reviewers will be specifically instructed to not penalize honesty concerning limitations.
    \end{itemize}

\item {\bf Theory Assumptions and Proofs}
    \item[] Question: For each theoretical result, does the paper provide the full set of assumptions and a complete (and correct) proof?
    \item[] Answer: \answerYes{} 
    \item[] Justification: We state in our \cref{sec:supralaplacian} that a connected graph has its second eigenvalue strictly positive. We include this proof and its source in Appendix \cref{app:supralap}.
    \item[] Guidelines:
    \begin{itemize}
        \item The answer NA means that the paper does not include theoretical results. 
        \item All the theorems, formulas, and proofs in the paper should be numbered and cross-referenced.
        \item All assumptions should be clearly stated or referenced in the statement of any theorems.
        \item The proofs can either appear in the main paper or the supplemental material, but if they appear in the supplemental material, the authors are encouraged to provide a short proof sketch to provide intuition. 
        \item Inversely, any informal proof provided in the core of the paper should be complemented by formal proofs provided in Appendix or supplemental material.
        \item Theorems and Lemmas that the proof relies upon should be properly referenced. 
    \end{itemize}

    \item {\bf Experimental Result Reproducibility}
    \item[] Question: Does the paper fully disclose all the information needed to reproduce the main experimental results of the paper to the extent that it affects the main claims and/or conclusions of the paper (regardless of whether the code and data are provided or not)?
    \item[] Answer: \answerYes{} 
    \item[] Justification: We provide a comprehensive overview of our model in \cref{fig:model}. Detailed discussions of our architecture can be found in \cref{sec:full_attention_transformer,sec:xa}. Also, algorithm of our supra-Laplacian computation is in \cref{app:supralap}. Implementation specifics are outlined in the implementation details section of \cref{sec:experiments} and further elaborated in \cref{app:param}.
    \item[] Guidelines:
    \begin{itemize}
        \item The answer NA means that the paper does not include experiments.
        \item If the paper includes experiments, a No answer to this question will not be perceived well by the reviewers: Making the paper reproducible is important, regardless of whether the code and data are provided or not.
        \item If the contribution is a dataset and/or model, the authors should describe the steps taken to make their results reproducible or verifiable. 
        \item Depending on the contribution, reproducibility can be accomplished in various ways. For example, if the contribution is a novel architecture, describing the architecture fully might suffice, or if the contribution is a specific model and empirical evaluation, it may be necessary to either make it possible for others to replicate the model with the same dataset, or provide access to the model. In general. releasing code and data is often one good way to accomplish this, but reproducibility can also be provided via detailed instructions for how to replicate the results, access to a hosted model (e.g., in the case of a large language model), releasing of a model checkpoint, or other means that are appropriate to the research performed.
        \item While NeurIPS does not require releasing code, the conference does require all submissions to provide some reasonable avenue for reproducibility, which may depend on the nature of the contribution. For example
        \begin{enumerate}
            \item If the contribution is primarily a new algorithm, the paper should make it clear how to reproduce that algorithm.
            \item If the contribution is primarily a new model architecture, the paper should describe the architecture clearly and fully.
            \item If the contribution is a new model (e.g., a large language model), then there should either be a way to access this model for reproducing the results or a way to reproduce the model (e.g., with an open-source dataset or instructions for how to construct the dataset).
            \item We recognize that reproducibility may be tricky in some cases, in which case authors are welcome to describe the particular way they provide for reproducibility. In the case of closed-source models, it may be that access to the model is limited in some way (e.g., to registered users), but it should be possible for other researchers to have some path to reproducing or verifying the results.
        \end{enumerate}
    \end{itemize}

\item {\bf Open access to data and code}
    \item[] Question: Does the paper provide open access to the data and code, with sufficient instructions to faithfully reproduce the main experimental results, as described in supplemental material?
    \item[] Answer: \answerYes{} 
    \item[] Justification: The code of SLATE is provided at this link: \href{https://github.com/ykrmm/SLATE}{https://github.com/ykrmm/SLATE}. Our code is designed to be comprehensible, ensuring that all presented results are reproducible.
    \item[] Guidelines:
    \begin{itemize}
        \item The answer NA means that paper does not include experiments requiring code.
        \item Please see the NeurIPS code and data submission guidelines (\url{https://nips.cc/public/guides/CodeSubmissionPolicy}) for more details.
        \item While we encourage the release of code and data, we understand that this might not be possible, so “No” is an acceptable answer. Papers cannot be rejected simply for not including code, unless this is central to the contribution (e.g., for a new open-source benchmark).
        \item The instructions should contain the exact command and environment needed to run to reproduce the results. See the NeurIPS code and data submission guidelines (\url{https://nips.cc/public/guides/CodeSubmissionPolicy}) for more details.
        \item The authors should provide instructions on data access and preparation, including how to access the raw data, preprocessed data, intermediate data, and generated data, etc.
        \item The authors should provide scripts to reproduce all experimental results for the new proposed method and baselines. If only a subset of experiments are reproducible, they should state which ones are omitted from the script and why.
        \item At submission time, to preserve anonymity, the authors should release anonymized versions (if applicable).
        \item Providing as much information as possible in supplemental material (appended to the paper) is recommended, but including URLs to data and code is permitted.
    \end{itemize}

\item {\bf Experimental Setting/Details}
    \item[] Question: Does the paper specify all the training and test details (e.g., data splits, hyperparameters, how they were chosen, type of optimizer, etc.) necessary to understand the results?
    \item[] Answer: \answerYes{} 
    \item[] Justification: The selection of datasets, their splitting, and descriptions are presented in \cref{app:dts}. We use the same evaluation protocols as papers well-recognized by the community \cite{yu2023towardslibrary,yang2021discrete-timespace}. Hyperparameter optimization is detailed in \cref{app:param}, and the optimizer settings are described at the beginning of \cref{sec:experiments}.
    \item[] Guidelines:
    \begin{itemize}
        \item The answer NA means that the paper does not include experiments.
        \item The experimental setting should be presented in the core of the paper to a level of detail that is necessary to appreciate the results and make sense of them.
        \item The full details can be provided either with the code, in appendix, or as supplemental material.
    \end{itemize}

\item {\bf Experiment Statistical Significance}
    \item[] Question: Does the paper report error bars suitably and correctly defined or other appropriate information about the statistical significance of the experiments?
    \item[] Answer: \answerYes{} 
    \item[] Justification: Following the protocols we are based on, all results in the paper, including those from ablation studies, are averaged over 5 runs with the standard deviation reported.

    \item[] Guidelines:
    \begin{itemize}
        \item The answer NA means that the paper does not include experiments.
        \item The authors should answer "Yes" if the results are accompanied by error bars, confidence intervals, or statistical significance tests, at least for the experiments that support the main claims of the paper.
        \item The factors of variability that the error bars are capturing should be clearly stated (for example, train/test split, initialization, random drawing of some parameter, or overall run with given experimental conditions).
        \item The method for calculating the error bars should be explained (closed form formula, call to a library function, bootstrap, etc.)
        \item The assumptions made should be given (e.g., Normally distributed errors).
        \item It should be clear whether the error bar is the standard deviation or the standard error of the mean.
        \item It is OK to report 1-sigma error bars, but one should state it. The authors should preferably report a 2-sigma error bar than state that they have a 96\% CI, if the hypothesis of Normality of errors is not verified.
        \item For asymmetric distributions, the authors should be careful not to show in tables or figures symmetric error bars that would yield results that are out of range (e.g. negative error rates).
        \item If error bars are reported in tables or plots, The authors should explain in the text how they were calculated and reference the corresponding figures or tables in the text.
    \end{itemize}

\item {\bf Experiments Compute Resources}
    \item[] Question: For each experiment, does the paper provide sufficient information on the computer resources (type of compute workers, memory, time of execution) needed to reproduce the experiments?
    \item[] Answer: \answerYes{} 
    \item[] Justification: The analysis of the time-memory efficiency of our model is presented in \cref{tab:efficiency}, where we also compare it with other state-of-the-art models. We also detailed the number of parameters of \ours. 
    \item[] Guidelines:
    \begin{itemize}
        \item The answer NA means that the paper does not include experiments.
        \item The paper should indicate the type of compute workers CPU or GPU, internal cluster, or cloud provider, including relevant memory and storage.
        \item The paper should provide the amount of compute required for each of the individual experimental runs as well as estimate the total compute. 
        \item The paper should disclose whether the full research project required more compute than the experiments reported in the paper (e.g., preliminary or failed experiments that didn't make it into the paper). 
    \end{itemize}
    
\item {\bf Code Of Ethics}
    \item[] Question: Does the research conducted in the paper conform, in every respect, with the NeurIPS Code of Ethics \url{https://neurips.cc/public/EthicsGuidelines}?
    \item[] Answer: \answerYes{} 
    \item[] Justification: Our research adheres strictly to the NeurIPS Code of Ethics. Our study does not involve sensitive data or unethical practices, and we have followed all relevant guidelines to ensure ethical compliance throughout our work.
    \item[] Guidelines:
    \begin{itemize}
        \item The answer NA means that the authors have not reviewed the NeurIPS Code of Ethics.
        \item If the authors answer No, they should explain the special circumstances that require a deviation from the Code of Ethics.
        \item The authors should make sure to preserve anonymity (e.g., if there is a special consideration due to laws or regulations in their jurisdiction).
    \end{itemize}

\item {\bf Broader Impacts}
    \item[] Question: Does the paper discuss both potential positive societal impacts and negative societal impacts of the work performed?
    \item[] Answer: \answerNA{} 
    \item[] Justification: Our model is a discriminative model for link prediction on academic datasets, which do not contain any private or sensitive information.
    \item[] Guidelines:
    \begin{itemize}
        \item The answer NA means that there is no societal impact of the work performed.
        \item If the authors answer NA or No, they should explain why their work has no societal impact or why the paper does not address societal impact.
        \item Examples of negative societal impacts include potential malicious or unintended uses (e.g., disinformation, generating fake profiles, surveillance), fairness considerations (e.g., deployment of technologies that could make decisions that unfairly impact specific groups), privacy considerations, and security considerations.
        \item The conference expects that many papers will be foundational research and not tied to particular applications, let alone deployments. However, if there is a direct path to any negative applications, the authors should point it out. For example, it is legitimate to point out that an improvement in the quality of generative models could be used to generate deepfakes for disinformation. On the other hand, it is not needed to point out that a generic algorithm for optimizing neural networks could enable people to train models that generate Deepfakes faster.
        \item The authors should consider possible harms that could arise when the technology is being used as intended and functioning correctly, harms that could arise when the technology is being used as intended but gives incorrect results, and harms following from (intentional or unintentional) misuse of the technology.
        \item If there are negative societal impacts, the authors could also discuss possible mitigation strategies (e.g., gated release of models, providing defenses in addition to attacks, mechanisms for monitoring misuse, mechanisms to monitor how a system learns from feedback over time, improving the efficiency and accessibility of ML).
    \end{itemize}
    
\item {\bf Safeguards}
    \item[] Question: Does the paper describe safeguards that have been put in place for responsible release of data or models that have a high risk for misuse (e.g., pretrained language models, image generators, or scraped datasets)?
    \item[] Answer: \answerNA{} 
    \item[] Justification: We don't release new data or harmful generative models. 
    \item[] Guidelines:
    \begin{itemize}
        \item The answer NA means that the paper poses no such risks.
        \item Released models that have a high risk for misuse or dual-use should be released with necessary safeguards to allow for controlled use of the model, for example by requiring that users adhere to usage guidelines or restrictions to access the model or implementing safety filters. 
        \item Datasets that have been scraped from the Internet could pose safety risks. The authors should describe how they avoided releasing unsafe images.
        \item We recognize that providing effective safeguards is challenging, and many papers do not require this, but we encourage authors to take this into account and make a best faith effort.
    \end{itemize}

\item {\bf Licenses for existing assets}
    \item[] Question: Are the creators or original owners of assets (e.g., code, data, models), used in the paper, properly credited and are the license and terms of use explicitly mentioned and properly respected?
    \item[] Answer: \answerYes{} 
    \item[] Justification: We properly cite all the datasets and baselines used in our paper. Each dataset and model is credited to its original creators, and we adhere to the specified licenses and terms of use. The evaluation protocols we employ are based on established standards from previous works, ensuring compliance with the original authors' conditions.
    \item[] Guidelines:
    \begin{itemize}
        \item The answer NA means that the paper does not use existing assets.
        \item The authors should cite the original paper that produced the code package or dataset.
        \item The authors should state which version of the asset is used and, if possible, include a URL.
        \item The name of the license (e.g., CC-BY 4.0) should be included for each asset.
        \item For scraped data from a particular source (e.g., website), the copyright and terms of service of that source should be provided.
        \item If assets are released, the license, copyright information, and terms of use in the package should be provided. For popular datasets, \url{paperswithcode.com/datasets} has curated licenses for some datasets. Their licensing guide can help determine the license of a dataset.
        \item For existing datasets that are re-packaged, both the original license and the license of the derived asset (if it has changed) should be provided.
        \item If this information is not available online, the authors are encouraged to reach out to the asset's creators.
    \end{itemize}

\item {\bf New Assets}
    \item[] Question: Are new assets introduced in the paper well documented and is the documentation provided alongside the assets?
    \item[] Answer: \answerYes{} 
    \item[] Justification: Our paper introduces new assets, including code implementations and datasets for DTDG. Detailed documentation is provided alongside these assets, following structured templates that include information about training and licensing. 
    \item[] Guidelines:
    \begin{itemize}
        \item The answer NA means that the paper does not release new assets.
        \item Researchers should communicate the details of the dataset/code/model as part of their submissions via structured templates. This includes details about training, license, limitations, etc. 
        \item The paper should discuss whether and how consent was obtained from people whose asset is used.
        \item At submission time, remember to anonymize your assets (if applicable). You can either create an anonymized URL or include an anonymized zip file.
    \end{itemize}

\item {\bf Crowdsourcing and Research with Human Subjects}
    \item[] Question: For crowdsourcing experiments and research with human subjects, does the paper include the full text of instructions given to participants and screenshots, if applicable, as well as details about compensation (if any)? 
    \item[] Answer: \answerNA{} 
    \item[] Justification: We do not conduct research with human subjects.
    \item[] Guidelines:
    \begin{itemize}
        \item The answer NA means that the paper does not involve crowdsourcing nor research with human subjects.
        \item Including this information in the supplemental material is fine, but if the main contribution of the paper involves human subjects, then as much detail as possible should be included in the main paper. 
        \item According to the NeurIPS Code of Ethics, workers involved in data collection, curation, or other labor should be paid at least the minimum wage in the country of the data collector. 
    \end{itemize}

\item {\bf Institutional Review Board (IRB) Approvals or Equivalent for Research with Human Subjects}
    \item[] Question: Does the paper describe potential risks incurred by study participants, whether such risks were disclosed to the subjects, and whether Institutional Review Board (IRB) approvals (or an equivalent approval/review based on the requirements of your country or institution) were obtained?
    \item[] Answer: \answerNA{} 
    \item[] Justification:  We do not conduct research with human subjects.
    \item[] Guidelines:
    \begin{itemize}
        \item The answer NA means that the paper does not involve crowdsourcing nor research with human subjects.
        \item Depending on the country in which research is conducted, IRB approval (or equivalent) may be required for any human subjects research. If you obtained IRB approval, you should clearly state this in the paper. 
        \item We recognize that the procedures for this may vary significantly between institutions and locations, and we expect authors to adhere to the NeurIPS Code of Ethics and the guidelines for their institution. 
        \item For initial submissions, do not include any information that would break anonymity (if applicable), such as the institution conducting the review.
    \end{itemize}

\end{enumerate}
\end{document}